\documentclass[11pt]{article}

\usepackage[preprint]{acl}

\usepackage{times}
\usepackage{latexsym}

\usepackage[T1]{fontenc}
\usepackage[utf8]{inputenc}

\usepackage{microtype}

\usepackage{inconsolata}
\usepackage{caption}
\usepackage{subcaption}
\usepackage{graphicx}
\usepackage[most]{tcolorbox}
\usepackage{fontawesome5}
\usepackage{xspace}
\usepackage{booktabs}
\lstdefinelanguage{json}{
    basicstyle=\small\ttfamily,
    numbers=left,
    numberstyle=\tiny\color{gray},
    stepnumber=1,
    numbersep=8pt,
    showstringspaces=false,
    breaklines=true,
    frame=lines,
    stringstyle=\color{blue},
    keywordstyle=\color{magenta},
}

\tcbuselibrary{breakable, skins}
\newcommand{\method}{APEX\xspace}
\newcommand{\bench}{APEX-Bench\xspace}

\title{APEX: Academic Poster Editing Agentic Expert}

\author{
  Chengxin Shi\thanks{\ \ Equal contribution.} \quad
  Qinnan Cai\footnotemark[1] \quad
  Zeyuan Chen\footnotemark[1] \quad
  Long Zeng\footnotemark[1] \\
  \textbf{Yibo Zhao} \quad
  \textbf{Jing Yu} \quad
  \textbf{Jianxiang Yu} \quad
\textbf{Xiang Li}\thanks{Corresponding Author: \texttt{xiangli@dase.ecnu.edu.cn}} \\  School of Data Science and Engineering, East China Normal University \\
  \texttt{chengxinshi@stu.ecnu.edu.cn} 
}
\begin{document}
\maketitle
% -----------------------
\begin{abstract}
Designing academic posters is a labor-intensive process requiring the precise balance of high-density content and sophisticated layout. While existing paper-to-poster generation methods automate initial drafting, they are typically single-pass and non-interactive, often fail to align with complex, subjective user intent. To bridge this gap, we propose \method (\underline{A}cademic \underline{P}oster \underline{E}diting agentic e\underline{X}pert), the first agentic framework for interactive academic poster editing, supporting fine-grained control with robust multi-level API-based editing and a review-and-adjustment Mechanism. 
% . Our approach features two core designs: (i) Robust Multi-level API-based Editing, which employs a hierarchical API suite to precisely manipulate dense element groups (e.g., sections); and (ii) a Review-and-Adjustment Mechanism that performs post-edit audits to mitigate "one-go" execution errors. 
% To support research in this emerging domain, 
In addition,
we introduce \bench, the first systematic benchmark comprising 514 academic poster editing instructions, categorized by a multi-dimensional taxonomy including operation type, difficulty, and abstraction level, constructed via reference-guided and reference-free strategies to ensure realism and diversity.
% . \bench is structured by a multi-dimensional taxonomy spanning operation type, paper-dependency, difficulty, and abstraction level. 
% Coupled with a VLM-as-a-judge evaluation protocol assessing fulfillment, scope, and harmony, experimental results show that \method significantly outperforms regeneration and code-based baselines, providing a reliable, intent-aligned solution for professional poster refinement.
We further establish a multi-dimensional VLM-as-a-judge evaluation protocol to assess instruction fulfillment, modification scope, and visual consistency \& harmony. Experimental results demonstrate that \method significantly outperforms baseline methods.
% existing
% regeneration and code-based 
% methods, 
% providing a more reliable and intent-aligned solution for professional academic poster editing.
Our implementation is available at \url{https://github.com/Breesiu/APEX}.
 % refinement.
\end{abstract}

\section{Introduction}

Academic posters are a critical medium for disseminating research ideas at conferences, requiring concise content selection, clear visual structure, and careful layout design under high information density. Due to the substantial manual effort required to design posters, recent studies~\cite{pang2025paperposter, sun2025p2pautomatedpapertopostergeneration, zhang2025postergen,gao2025postermakerhighqualityproductposter} on paper-to-poster generation aim to automatically distill long-form academic papers into visually structured posters, often leveraging large language models (LLMs) and multi-agent frameworks to extract key content and generate layouts or rendering code (e.g., HTML/CSS or PPT-based representations).

% [引出研究的问题，以及研究的必要性] 
However, these existing paper-to-poster methods treat the task as a \emph{single-pass}, \emph{non-interactive} generation problem, where a complete poster is produced in one step. This formulation is misaligned with real-world poster design workflows for two reasons: \textbf{(i)} Due to inherent limitations in current model capabilities, posters generated by existing paper-to-poster agents often exhibit substantial discrepancies from user-designed posters~\cite{pang2025paperposter}. As a result, significant and non-trivial modifications are still required, highlighting the necessity of leveraging LLMs for subsequent iterative editing and refinement.
\textbf{(ii)} User requirements for poster design are often complex
% \scx{evidence} 
and subjective, as poster design itself is a highly intricate task. Consequently, it is difficult for users to clearly and comprehensively articulate all their intentions within a single round of interaction.

% [现存方法的问题]
% To address this limitation, one natural solution is to leverage advanced multi-modal foundation models (e.g., GPT or Gemini) to support interactive poster editing using both natural language instructions and paper content as inputs. Such models can respond flexibly to user requests, but when applied to interactive poster editing, they typically regenerate the entire poster after each user interaction. In practice, this regenerate-all paradigm introduces two major issues.
Several paradigms can be considered to address this limitation, which broadly fall into two categories. \textbf{(i) Regeneration-based approaches} are exemplified by the Regenerate-All strategy, which leverages multi-modal foundation models (e.g., Gemini) to reconstruct the entire poster at each interaction. For example, upon each modification request, the system feeds the original poster image together with the user’s text instruction into a multi-modal generative model, and regenerates the entire poster. However, due to inherent model hallucinations and the nature of global rewriting, such approaches often introduce erroneous modifications or unnecessary changes beyond the user’s intended scope. \textbf{(ii) Generic slide editing approaches} perform slide editing on academic poster, as many of them are authored in slide formats (e.g., \texttt{.pptx}). 
% instead seek to modify posters programmatically. 
One line of work prompts LLMs to directly generate editing scripts, but these scripts frequently suffer from high execution error rates and unsatisfactory editing quality. Another line of work employs generic slide-editing agents~\cite{jung2025talk, guo2024pptc} to manipulate posters.
% as many academic posters are authored in slide formats (e.g., \texttt{.pptx}). 
However, these generic slide-editing agents lack the domain-specific understanding required for academic poster design, which demands comprehension of the underlying paper content as well as tightly coupled layout and visual element organization.
% While API-driven frameworks for general slide editing like PPTC, they lack multimodal support for paper-based content updates and rely on low-level APIs that fail to capture the high-level semantic requirements of academic posters.

% [解决问题的方法]
% First, due to model hallucinations and global rewriting, regenerated posters often contain unnecessary or even erroneous modifications beyond the user’s intended scope, which hinders precise and localized edits. 
To address these issues,
in this paper, we propose \method, the first academic poster editing agent framework, which is built upon two core designs. \textbf{(i) Robust multi-level API-based editing.} Unlike regeneration-based strategies, which are prone to global hallucinations, we formulate poster editing as a sequence of operations, where each operation is mapped to a specific API. This design enables
% consistent and 
localized modifications, supporting fine-grained control while avoiding unintended global changes. Compared to generic slide
% code-based 
editing approaches, our design is better suited for complex academic poster editing scenarios involving numerous, diverse and densely arranged element groups (e.g. sections). This advantage stems from our multi-level API design, which includes both low-level APIs (e.g., insert a single image) and high-level APIs that enable group-level and multi-attribute operations. In addition, we introduce a robust fault-tolerant execution mechanism for API sequences, ensuring that failures at intermediate steps do not invalidate the entire editing process.
% As a result, our approach enables robust execution and achieves a significantly lower execution error rate than code-based methods.
% \scx{"only targested elements are modified oriented.." need delete. --to complexity of academoic poster, 3-4slide, layout, isolation} \textbf{(ii) Iterative execution and correction.}\scx{uncomment}
\textbf{(ii) Review-and-adjustment mechanism.}
Beyond basic planning and execution of API sequences, our agent framework incorporates an additional review-and-adjustment module. This module mitigates the impact of instruction-following deficiencies in academic poster editing that stem from insufficient domain-specific capabilities of the underlying foundation models, which are prevalent in both regeneration-based and generic slide editing approaches.
% \scx{seems all methods have this problem}. 
Specifically, the system reviews the edited poster, focusing on the modified elements, to verify instruction fulfillment and detect redundant or incorrect modifications, triggering adjustments when necessary. 

Further, to systematically evaluate academic poster editing, we introduce \bench, the first comprehensive benchmark for this emerging task. Our benchmark comprises 514 poster editing instructions associated with 59 paper–poster pairs, collected from top-tier international AI conferences between 2023 and 2025, including ICLR, ICML, and NeurIPS. To ensure that these instructions reflect realistic academic poster editing scenarios, we adopt a rigorous and systematic instruction design process. In particular, we treat AI-generated academic posters as initial drafts and synthesize editing instructions using Vision–Language Models (VLMs) \cite{yin2024survey} through two complementary strategies: \textbf{(i) reference-guided synthesis}, which uses author-designed posters as references and derives instructions by analyzing differences between AI-generated and human-authored posters; and \textbf{(ii) reference-free synthesis}, which directly prompts VLMs to propose improvement instructions based on content, structure, and aesthetics, serving as a complementary source to enhance instruction diversity. All synthesized instructions are subsequently reviewed and refined through human verification to ensure quality and correctness. We further introduce a multi-dimensional taxonomy that categorizes instructions by operation type, paper relevance, difficulty level, and abstraction level, to support fine-grained evaluation.

To accompany our benchmark dataset, we propose a VLM-as-a-judge~\cite{chen2024mllm, zheng2023judging} evaluation protocol that assesses poster quality under a given editing instruction from three aspects: \textbf{(i) instruction fulfillment}, which measures how completely the instruction is fulfilled and whether the modified or newly integrated content remains faithful to the source paper; \textbf{(ii) modification scope}, which evaluates whether unnecessary or unintended changes are introduced beyond the intended editing scope; and \textbf{(iii) visual consistency \& harmony}, which assesses whether the newly integrated or modified elements are visually coherent and natural within the overall design (e.g., layout and style). Collectively, these criteria enable a comprehensive evaluation of instruction compliance, editing scope control, and visual coherence.
% \scx{construction process} 
% Our poster cover 4 instruction types including xx, xx, xx and xx, ensuring fine-grained blabla.
% To accompany our benchmark dataset, we propose an LLM-as-a-judge evaluation protocol to assess poster quality under a given editing instruction from three complementary aspects. \textbf{(i) Instruction fulfillment}, which measures whether the user instructions are correctly and completely executed without omissions. \textbf{(ii) Modification scope}, which evaluates whether the model introduces unnecessary or erroneous changes beyond the intended scope. \textbf{(iii) Visual consistency}\scx{..}, which assesses whether the poster layout remains coherent and free from issues such as element overlap or occlusion. This is particularly important when user instructions are vague or ambiguous, where a literal fulfillment of the instruction could still result in visually unreasonable outcomes. For example, the instruction may ask to slightly enlarge a title, and the model follows it, yet the enlarged title occludes an image.
Overall, the main contributions of this work are summarized as follows:
% \begin{itemize}

\noindent{\small$\bullet$}  To our best, we present the first systematic study of interactive academic poster editing.
% moving beyond single-pass paper-to-poster generation.

\noindent{\small$\bullet$} We propose \method, an agent-based framework that enables interactive and fine-grained academic poster editing.

\noindent{\small$\bullet$} We introduce \bench, the first comprehensive benchmark for academic poster editing.
% together with an VLM-as-a-judge evaluation protocol that assesses editing quality from multiple critical dimensions.

\noindent{\small$\bullet$} Experimental results show that our approach significantly outperforms competitors, producing more reliable and intent-aligned poster edits.

% \end{itemize}

\section{Related Work}

\textbf{Academic poster generation.} Transforming full-length scientific papers into visually structured posters is an important and emerging research direction, as it requires joint reasoning over long-context text, figures, and layout. P2P~\cite{sun2025p2pautomatedpapertopostergeneration} and Paper2Poster~\cite{pang2025paperposter} are among the earliest representative works in this area, both introducing the first batch of benchmarks for paper-to-poster generation and addressing the task with multi-agent frameworks. Building on this line of work, PosterGen~\cite{zhang2025postergen} further advances the field by explicitly incorporating aesthetic design principles such as narrative structure, layout balance, color harmony, and typography into specialized agents, thereby improving visual quality and readability. 
Further, from a structural perspective, PosterForest~\cite{choi2025posterforest} introduces a hierarchical Poster Tree representation that jointly encodes document structure and visual–textual relationships, enabling multi-agent collaboration between content and layout agents to iteratively refine logical consistency and visual coherence. However, 
these existing paper-to-poster methods primarily focus on fully automated, single-pass generation, producing a complete poster in one step without iterative user feedback, which limits their ability to correct errors or flexibly adjust content and layout in practice. This limitation highlights the need for interactive poster generation, where users can iteratively guide and refine the output through direct interaction with the model.

\textbf{Agent-based slide editing.} Research on automatic slide editing provides useful insights for interactive poster editing, as most poster generation methods output PPTX slides. Existing benchmarks such as PPTC~\cite{guo2024pptc}, PPTC-R~\cite{zhang2024pptc}, PPTBench~\cite{huang2025pptbenchholisticevaluationlarge} and PPTArena~\cite{2025pptarena} define a range of slide editing tasks and operations. However, these benchmarks primarily focus on generic slide-level edits and do not evaluate quality in the academic poster domain, where correctness depends on understanding and preserving the semantic structure of the source academic paper, rather than executing generic slide-level operations. Most existing slide editing approaches adopt agent-based frameworks, such as Talk-to-Your-Slides~\cite{jung2025talk} and AUTO-SLIDES~\cite{yang2025autoslidesinteractivemultiagent}. 
Despite the effectiveness for basic slide operations, these systems are not specifically designed for academic posters, whose editing capability is thus restricted.
% which require paper-aware understanding to determine semantically appropriate edits across multiple sections, a capability not supported by standard slide editing frameworks. 
Moreover, these works rely exclusively on textual instructions and lack explicit feedback and refinement mechanisms. Recent work on layout reasoning~\cite{gao2025layoutcotunleashingdeep,guerreiro2024layoutflow,shen2025layoutrectifier} and concurrent slide editing research~\cite{2025pptarena,jung2025talk,yun2025designlab} demonstrates that
% iterative  
review and refinement play a critical role in improving instruction-following accuracy and layout quality. These findings further highlight the importance of incorporating feedback-driven refinement for layout-centric tasks requiring precise instruction following in academic poster editing.

\section{Problem Definition}
% Given an academic poster $\mathcal{P}_{src}$ (in .pptx format), a source paper $\mathcal{D}$, and a user instruction $\mathcal{I}$ such as content revision, layout reconfiguration. the objective of the automated system is to edit $\mathcal{P}_{src}$ to fulfill $\mathcal{I}$ accurately, efficiently, and affordably, then output edited poster $P_{tgt}$(Editable .pptx format) to enable interactive human-in-the-loop refinement process. The output edited poster should satisfy requested changes while strictly preserving the integrity of unrequested sections, maintaining the visual consistency \& harmony of modified elements if user don't specify attribute of element.

The task of interactive academic poster editing concerns revising an existing poster according to natural language instructions while preserving editability for subsequent human-in-the-loop refinement.

Let $\mathcal{P}$, $\mathcal{M}$ and $\mathcal{I}$ denote source poster set, source paper set and user instruction set, respectively.
Formally, given a source academic poster \( P_{\text{src}} \in \mathcal{P} \) in an editable format (e.g., \texttt{.pptx}), a corresponding source paper \( M \in \mathcal{M} \), and a user instruction \( I \in \mathcal{I} \) (e.g., content revision or layout reconfiguration), the goal of automated poster editing is to generate an edited target poster \( P_{\text{tgt}} \) in an editable format that satisfies the user’s editing needs.

We model the poster editing process as a parameterized editing policy:
\begin{equation}
P_{\text{tgt}} = \pi_{\theta}(P_{\text{src}}, M, I),
\end{equation}
where \( \pi_{\theta} \) denotes an editing policy with parameters \( \theta \) that maps the source poster, the paper content, and the user instruction to an edited poster.

The quality of the edited poster is evaluated by a parameterized scoring function \( S_{\pi_{\phi}}(\cdot) \), where the score is induced by an evaluation policy \( \pi_{\phi} \) with parameters \( \phi \). This scoring function assigns higher values to edits that better align with user intent while avoiding unnecessary or unintended modifications. The objective of the system is to generate an edited poster that maximizes this quality score:
\begin{equation}
\max_{\theta} \; S_{\pi_{\phi}}(P_{\text{src}}, P_{\text{tgt}}, I, M).
\end{equation}

Accordingly, the task requires a well-defined editing policy \( \pi_{\theta} \), an evaluation policy \( \pi_{\phi} \) together with its induced scoring function \( S_{\pi_{\phi}} \), as well as a collection of editing instances \( (P_{\text{src}}, M, I) \) for validation and evaluation.
% This task has not been systematically studied in prior work for this task.

\begin{figure*}[!ht]
    \centering
    \includegraphics[width=1\linewidth]{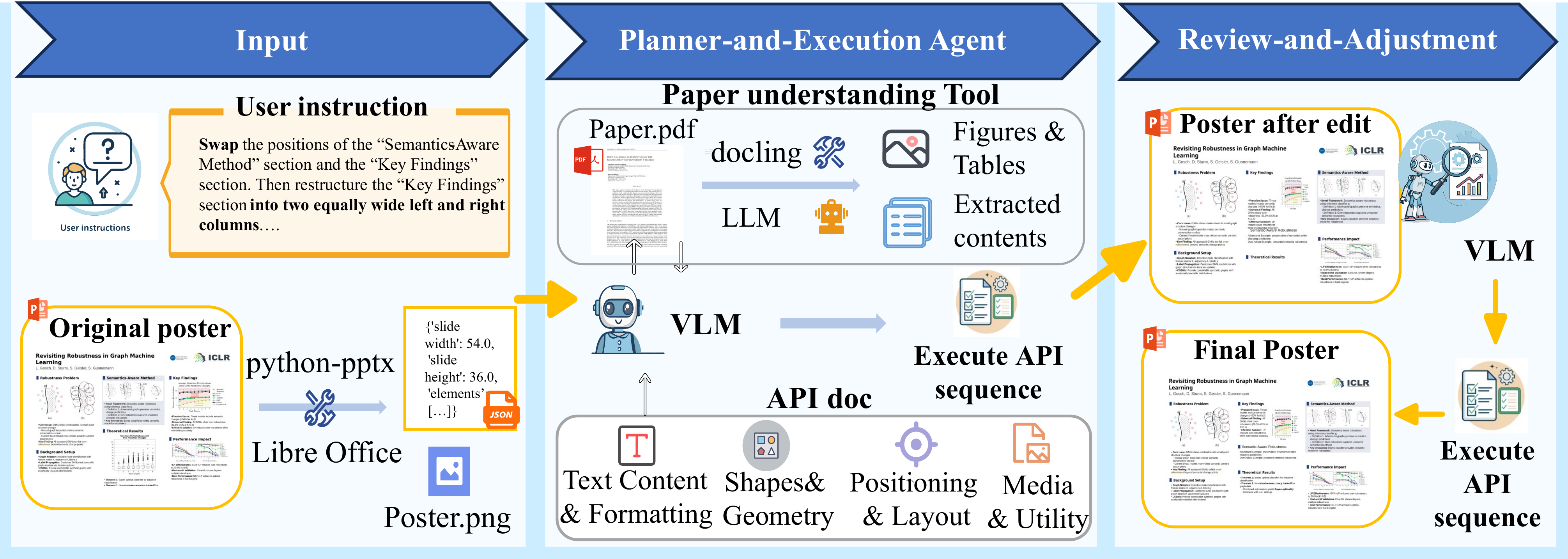}
    \caption{\textbf{Overview of the Multi-Agent Poster Editing Pipeline.} The framework comprises three collaborative stages: 
    (1) \textbf{Semantic Parsing and Element Serialization}: The source poster ${P}_\text{src}$ is parsed into structured JSON data $P^J_\text{src}$, extracting all element attributes for fine-grained control. 
    (2) \textbf{Planning and Execution}: A centralized agent synthesizes user instructions $I$, visual representation $P^V_\text{src}$, and JSON data $P^J_\text{src}$ to generate an execution plan. It optionally invokes a paper understanding tool to extract content from the paper $M$ before calling multi-level APIs to modify the \texttt{.pptx} file. 
    (3) \textbf{Review and Adjustment}: A quality assurance agent evaluates the edited visual output $P^V_\text{edited}$ against the original $P^V_\text{src}$ and the user's instruction $I$, specifically verifying the modified elements, and performs adjustments through additional API calls to ensure visual fidelity and instruction compliance.}
    % \scx{data flow is not clear, font size are too small?}
    \label{fig:pipeline_architecture}
\end{figure*}

\section{\method}  
%  is _pi \theta policy needed?
% In this section, we present \method,
% % \textbf{\method},
% a multi-agent framework for academic paper editing.
% % implemented as a precise academic poster editing policy $\pi_\theta$.
% % $\pi_\theta$, a multi-agent framework tailored for precise academic poster editing. 
% By decomposing complex poster editing process
% % revisions 
% into multi-level API operations and incorporating a review-and-adjustment mechanim, \method
% % \textbf{\method} 
% ensures high instruction adherence and visual harmony, while preserving the integrity of the original poster. The system's architecture is detailed below, beginning with its core design followed by the multi-agent workflow. 

\subsection{Core Design}
To achieve robust and aesthetically pleasing editing, our system employs two core strategies: Robust multi-level API-based editing and Review-and-Adjustment mechanism. These designs are specifically engineered to enhance the system's capacity for manipulating high-density elements in academic poster while mitigating the risks of ``one-go'' editing deviations in scenarios such as complex layout edit.
% By bridging the gap between abstract user intent and precise element control, they ensure that the system can perform stable and fine-grained modifications even under intricate editing requirements.
% These strategies are specifically designed to overcome the limitations of traditional slide editing methods\zl{these two problems only occur in slide editing?}, which often lack the granularity required for complex academic posters and are prone to fragility when handling advanced instruction, i.e. layout adjustments. 需要将各自的优点与方法对应起来？

\paragraph{Robust multi-level API-based editing.} We propose a robust multi-level API architecture designed to handle the high-density information contained in academic posters.
% \zl{confusing. Modification: } 
While retaining low-level APIs (e.g., moving single element, setting single attribute)
% \zl{what is low-level api? Modification: } 
for fine-grained control, we develop a suite of high-level APIs to manipulate the complex and numerous element group (e.g., section) in academic poster. For instance, \verb|move_group()| and \texttt{batch\_delete\_elements()} APIs falicitate edit on group-level element, while \texttt{text\_format\_brush()} enables batch update multiple attributes (e.g., font size, color, bolding, etc.). Further, different from executing all edits in one single step,
% all codes in one time, 
our system processes API calls
% processes each API call 
individually to ensure fault tolerance. 
% If an API fail to execute, the system will continue executing the remaining APIs, preventing a single failure from causing the entire editing to fail.
When an API call fails, the system continues executing the remaining operations, preventing a single failure from invalidating the entire editing process.
% Conversely, a single line error in code-based methods will lead to code execution interruption.
Conversely, if generate editing script to edit poster, a sinle line error will leads to code execution interruption.
% to mitigate system fragility\zl{why is the system fragile?}, these APIs are executed in isolation. This step-by-step atomic execution ensures that an error in a single API calling does not compromise the entire editing process, avoiding causing the entire editing failure.

% \paragraph{Robust multi-level API-based editing.} We propose a multi-level API architecture tailored to handle the high information density of academic posters. This architecture integrates high-level semantic abstractions with a robust execution mechanism. Specifically, beyond basic primitive operations (e.g., moving a single element or modifying an individual attribute), we developed a suite of high-level APIs designed for manipulating complex element hierarchies within PowerPoint files. For instance, APIs such as \verb|move_group()| and \texttt{batch\_delete\_elements()} facilitate group-level editing, while \texttt{text\_format\_brush()} enables the batch application of stylistic attributes (e.g., font size, color, and weight). Crucially, unlike monolithic execution paradigms, our system executes API calls atomically. This ensures fault tolerance: if a specific API call fails, it is bypassed, allowing the remaining operations to proceed without compromising the entire editing process.

\paragraph{Review-and-Adjustment Mechanism.} 
% We devise a multi-modal review-and-adjustment mechanism that acts as a quality assurance layer to guarantee visual quality and instruction fidelity. This mechanism is essential for two main reasons: first, sometimes the system need to rearrange many elements associated with instruction-related parts, which is hard to be done in one go; second, small mistakes or omissions in modification parts may lead to significant errors like overlap, misalignment, or visual inconsistency in the overall poster.
We devise a multi-modal review-and-adjustment mechanism that functions as a quality assurance layer to rectify potential errors from the initial editing phase. Specifically, when initial edits are completed, this mechanism reviews the edited poster
% focusing on the modified elements,
% and modified elements 
to validate instruction fulfillment and identify redundant and incorrect modification. If the initial edits fall short of the requirement, an adjustment phase will be triggered to refine edited poster further; otherwise, the poster is finalized, mitigating unnecessary manipulations. Moreover, this mechanism prioritizes the modification parts that are made in the initial editing phase, ensuring the whole process to be 
% efficient 
effective
and focused, 
avoiding
% iterative\zl{why}, 
full-scale poster review and modification.
% Then it determines whether to adjust edited poster further based on the accuracy of the initial modifications, ensuring. If necessary, an adjustment phase is triggered; otherwise, the poster is finalized. Furthermore, this mechanism prioritizes the modification parts that are made in the initial editing phase, ensuring the whole process to be efficient and focused, avoiding iterative, full-scale poster review and modification.\zl{put more attention on xxx, ensuring xxx, avoiding xxx}
% design the review-and-adjustment process to be efficient and focused, 

\begin{figure*}[t] 
    \centering
    \includegraphics[width=\linewidth]{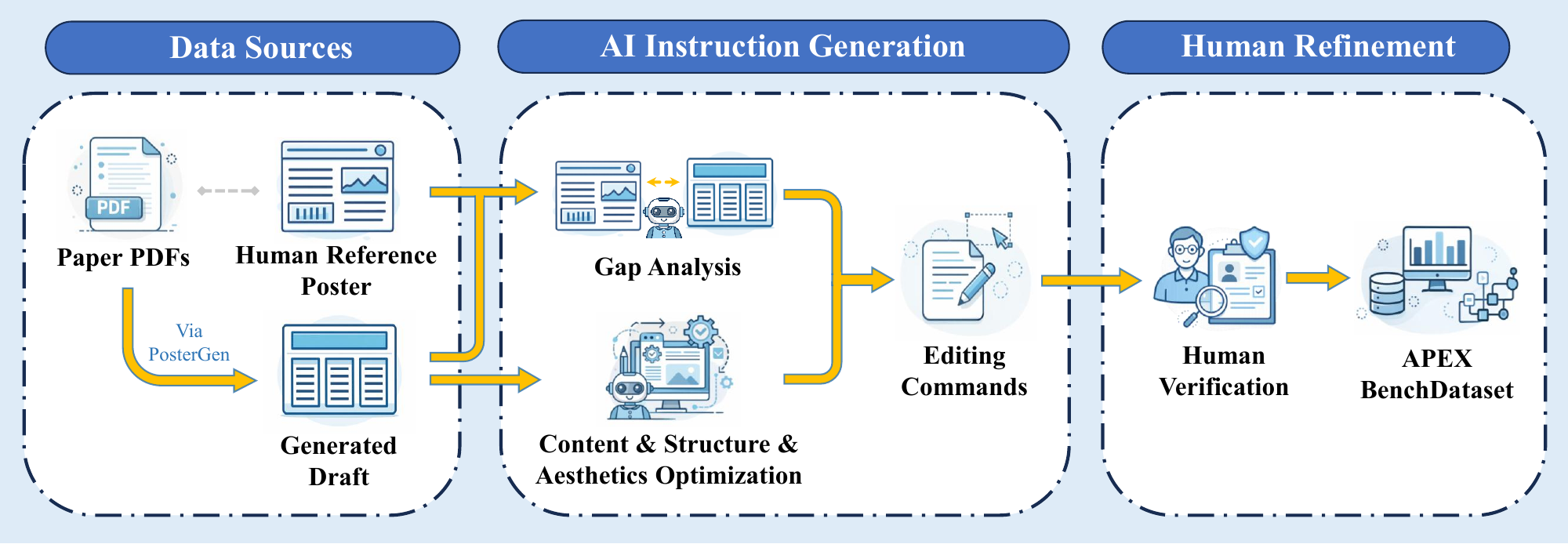}
    \caption{The data construction pipeline of \bench. Adopting a ``Model-assisted, Human-refined'' strategy, the workflow consists of three phases: (1) \textbf{Data Sources Preparation}, where initial drafts are synthesized from source papers via PosterGen; (2) \textbf{AI Instruction Generation}, where an VLM performs gap analysis and aesthetic optimization to derive preliminary editing commands; and (3) \textbf{Human Refinement}, where experts verify and adjust instructions to ensure feasibility and high quality.}
    \label{fig:pipeline}
\end{figure*}

\subsection{Overall Framework}

Our system operates as a multi-agent pipeline with three distinct stages,
% Our system operates as a multi-agent pipeline designed to parse, plan, execute, review and adjust inital poster edits, 
as shown in Figure \ref{fig:pipeline_architecture}. 
% The workflow transforms the raw poster input into structured data, orchestrates complex editing actions, and ensures edit quality through review-and-adjustment mechanism. 
% The framework consists of three distinct stages:

\paragraph{Semantic Parsing and Element Serialization.} The process commences by transforming the raw \texttt{.pptx} file into a structured, manipulable format to enable fine-grained observation and control. 
% using docling to parse figures and tables in paper $M$. 
Using \texttt{python-pptx} library, the system parses the source poster $P_{\text{src}}$ to extract attributes (e.g., IDs, text content, font sizes, positions, colors, borders, images, etc.) for all elements and serializes them into JSON data ${P}^J_\text{src}$, providing fine-grained information about each element for following editing steps to reference and modify specific elements accurately.

% \paragraph{Planning and Execution.} This stage serves as the system's central engine for planning and execution. The 
% \textit{planner-and-execution agent}, 
% % Planner-and-Execution Agent, 
% powered by a VLM, synthesizes multimodal inputs to orchestrate the editing workflow. It begins by analyzing user's instructions $I$, the parsed JSON data $P^J_{\text{src}}$, and the source poster image $P^V_{\text{src}}$, then determines whether to invoke the 
% % \zl{how to determine} whether to invoke the 
% \textit{paper understanding tool}.
% % \textit{Paper Understanding Tool}.
% This tool utilizes another VLM to extract relevant content from the source paper $M$ and provides figure/table paths and metadata (e.g., aspect ratios) pre-processed by the static parsing tool \texttt{Docling}.
% % for integration into the poster later. 
% If paper content is required, the agent will query this tool for the necessary information, and then generates the plan and a sequence of high-level API calls based on the provided API document. Otherwise, the planner directly generates a plan and an API sequence. These API calls are executed sequentially by the execution program to modify the poster $P_\text{src}$ and generate the updated version $P_\text{edited}$ with corresponding image $P^V_\text{edited}$.

\paragraph{Planning and Execution.} This stage serves as the system's central engine for planning and execution. The 
\textit{planner-and-execution agent}, 
% Planner-and-Execution Agent, 
powered by a VLM, synthesizes multimodal inputs to orchestrate the editing workflow. It begins by analyzing user's instructions $I$, the parsed JSON data $P^J_{\text{src}}$, and the source poster image $P^V_{\text{src}}$.
Based on this analysis, the agent determines whether the instruction requires content from the associated paper.
% then determines whether to invoke the 
% \zl{how to determine} whether to invoke the 
% \textit{paper understanding tool} depending on whether the user's instruction requires auxiliary paper content. 
% If such content is required, 
If paper content is required,
the agent invokes the \textit{paper understanding tool}
% queries this tool,
% where another VLM is utilized 
to extract relevant content from the source paper $M$
using a VLM, together with 
% alongside 
metadata (e.g., paths, aspect ratios) of figures and tables pre-processed by the static parsing toolkit \texttt{Docling}.
% and then generates 
Using these information and provided API document, the agent then generates
the plan and a sequence of 
% high-level 
multi-level
API calls. 
% based on the provided API document. 
Otherwise, 
when no paper content is needed, 
the planner directly generates a plan and an API sequence
without relying on the associated paper.
These API calls are executed sequentially by the execution program to modify the poster $P_\text{src}$, resulting in the updated version $P_\text{edited}$ with corresponding image $P^V_\text{edited}$.

\paragraph{Review and Adjustment.} The final stage is designed to guarantee visual fidelity and instruction adherence. The review-and-adjustment agent powered by a VLM reviews the initially edited poster image $P^V_\text{edited}$ by cross-referencing it with original poster image $P^V_\text{src}$, the executed API sequence, user's instruction $I$ and the JSON representation of original and modifed elements. This process aims to identify any mistakes, omissions or redundant modifications. If errors or unnecessary modifications are notable, this agent will generate a corrective API sequence to rectify these issues, thereby producing the finalized poster $P_\text{tgt}$.

\section{\bench}

\subsection{Data Construction}
To comprehensively evaluate model performance in academic poster editing, we curate \bench, 
a dataset comprising 514 editing instructions $\mathcal{I}$ derived from 59 high-quality papers $\mathcal{M}$. Each instruction $I \in \mathcal{I}$ is associated with a paper $M \in \mathcal{M}$, a human-authored reference poster $P_{\text{ref}}$, and an AI-generated initial poster $P_{\text{src}}$. Formally, the dataset is represented as $\mathcal{D}=\{(M, P_{\text{src}}, P_{\text{ref}}, I)\}$.
We employ a systematic ``model-assisted, human-refined'' strategy to generate these instructions, ensuring the coverage of diverse complexities and cross-modal dependencies. The construction pipeline, as illustrated in Figure~\ref{fig:pipeline}, consists of three phases:

\paragraph{Data Preparation and Initialization.} 
To cover diverse academic styles and topics, we curate a dataset $\mathcal{M}$ consisting of 59 source papers published between 2023 and 2025 in the top-tier AI conferences ICLR, ICML and NeurIPS. 
For each paper in $\mathcal{M}$, we utilize PosterGen~\cite{zhang2025postergen} to synthesize an initial poster draft $P_{\text{src}}$. This tool employs GLM-4.5V~\cite{vteam2026glm45vglm41vthinkingversatilemultimodal} and Gemini-3-Flash-Preview to parse paper content and generate layouts. We select PosterGen primarily due to its ability to output native, editable \texttt{.pptx} files and its hierarchical layout generation mechanism, which yields well-aligned and coherent layouts, providing a robust and structured starting point.

\paragraph{VLM-Driven Analysis and Optimization.}
To derive preliminary editing commands, we employ the VLM Gemini-3-Flash-Preview via a dual-branch prompting strategy designed to balance realism and diversity.
First, to ensure realism, we conduct a \textit{reference-guided gap analysis}, where the VLM compares the draft $P_\text{src}$ against the human-authored reference $P_\text{ref}$ to identify critical discrepancies, such as missing key figures, over-simplified text, and logical incoherence. However, relying solely on fixed references constrains the range of potential edits. Therefore, to enhance diversity, we further utilize a \textit{reference-free optimization} as a supplementary approach. In this independent analysis, the model evaluates $P_\text{src}$ to propose additional instructions for optimizing content, structure, and aesthetics that are not bounded by the reference, thereby enriching the dataset with a wider variety of editing intents. This dual-path strategy ensures comprehensive coverage of both content revisions and visual improvements.

\paragraph{Expert Verification and Refinement.} 
To ensure the high quality of our dataset, we recruit a team of five expert annotators to conduct a rigorous human review. Details regarding recruitment and annotation procedures are provided in Appendix~\ref{sec:annotation_process}. We adopt a review mechanism where each instruction is independently reviewed by at least two experts, with discrepancies resolved through discussion. The verification process strictly follows two phases: \textbf{(i) Filtering based on Validity and Professionalism:} Experts first inspect instructions for factual and academic standards. Instructions are discarded if they exhibit \textit{Validity Issues} (e.g., hallucinations or factual errors) or violate \textit{Professional Rationality} (e.g., aesthetically degrading or non-academic requests).
\textbf{(ii) Refining for Layout Feasibility:} For instructions that are valid in intent but spatially conflicted (e.g., a request to ``insert a figure'' into a fully occupied region), experts rewrite them into precise, multi-step instructions involving necessary layout adjustments. This ensures that \bench consists of logically complete and executable editing tasks.

\subsection{Dataset Statistics and Analysis}%Dataset分类学
We establish a multi-dimensional taxonomy to facilitate fine-grained evaluation. Detailed definitions and examples are provided in Appendix~\ref{sec:detailed_taxonomy}.
 \paragraph{Operation Category.} To systematically characterize the editing requirements of information-dense academic posters, we classify instructions into four primary categories. Text-related operations are the most prevalent ($79.77\%$), followed by overall layout ($59.34\%$), image adjustments ($47.67\%$), and shapes and elements ($33.27\%$). 
 Notably, the cumulative percentage exceeds $100\%$ because real-world editing instructions are composite. Unlike operations in generic slide editing, academic poster revision often requires simultaneous multi-aspect manipulation. For example, the instruction ``\emph{Move the image in the Method section to the left side and rearrange the text on the right}'' involves three dimensions: image adjustment, text rearrangement, and layout optimization.
 
 \paragraph{Difficulty Levels.} We classify instructions into four levels: Low, Medium, High, and Very High, based on two dimensions:
 \textbf{(i) Structural Complexity \& Scope}, which measures the granularity and extent of the required modifications, ranging from localized attribute updates (e.g., ``\emph{Adjust font styles}'') to complex global layout reconfigurations requiring multi-group coordination, and \textbf{(ii) Semantic Understanding \& Reasoning Depth}, which evaluates the cognitive demand to interpret and execute editing commands. 
 This spans from straightforward, explicit operations (e.g., ``\emph{Center align the title}'') to handling complex scenarios involving abstract instructions or paper-dependent reasoning, where the agent has to extract semantic contents from the paper and perform multi-step reasoning to derive the optimal editing strategy.

 \paragraph{Paper Related.} A salient feature of \bench is paper-related instructions, which accounts for 57.2\% of instructions. 
 Unlike generic editing tasks solvable via visual cues alone, this category mandates deep cross-modal reasoning, requiring agents to align editing decisions with the semantic context and factual details of the source paper.

\begin{table*}[!ht]
    \centering
    \scalebox{0.9}{
    \begin{tabular}{lcccccc}
    \hline
    \textbf{Methods} & \textbf{Error Rate ($\downarrow$)} & \textbf{I.F. ($\uparrow$)} & \textbf{M.S. ($\uparrow$)} & \textbf{V.C. ($\uparrow$)} & \textbf{Cost (\$)} \\ 
    \hline
    Direct Image Generation       & -               & \underline{6.48}           & 3.22           & 2.78           & 0.248 \\ 
    XML Generation      & \underline{0.39\%}  & 5.47           & 4.77           & \underline{3.72} & 0.051 \\ 
    Direct Script-based Editing         & 34.05\%         & 4.90           & \textbf{9.20}  & 4.21           & 0.012 \\ 
    PPTC                & \textbf{0.00\%} & 0.63           & 5.05           & 0.45           & 0.002 \\ 
    Talk-to-Your-Slides & 3.50\%          & 2.37           & 6.83           & 2.22           & 0.022 \\ 
    \textbf{\method}       & \textbf{0.00\%} & \textbf{7.97}  & \underline{9.04} & \textbf{7.18}  & 0.022 \\ 
    \hline
    \end{tabular}
    }
    \caption{Performance comparison on \bench. \textbf{I.F.}: Instruction Fulfillment, \textbf{M.S.}: Modification Scope, \textbf{V.C.}: Visual Consistency. Cost denotes the average dollar cost per edit.}
    \label{tab:main_results}
\end{table*}

\subsection{Evaluation}
Given the complex and visual nature of academic posters, we leverage a VLM-as-a-judge paradigm to evaluate the performance of the editing system. Specifically, the evaluation policy $\pi_{\phi}$ (implemented via a VLM) is provided with the target poster image $P^V_{\text{tgt}}$, the source poster image $P^V_{\text{src}}$, the user instruction $I$, and the source paper $M$. The evaluation is conducted across three distinct dimensions—\emph{Instruction Fulfillment}, \emph{Modification Scope}, and \emph{Visual Consistency \& Harmony}—to ensure a comprehensive assessment of the editing quality $S_{\pi_{\phi}}(P^V_{\text{src}}, P^V_{\text{tgt}}, I, M)$. Each dimension has its own evaluation checklist and scoring rubric (see Appendix~\ref{sec:appendix-protocol}) to ensure a comprehensive assessment of the editing quality. The detailed evaluation metrics are given as follows: \textbf{(i) Instruction Fulfillment.} This metric evaluates the edited poster’s compliance with specific user instructions as well as the factual integrity of content integrated from the source paper. 
% This metric assesses the fidelity of the editing policy $\pi_{\theta}$ in executing the user's specific requests. It verifies whether the target poster $P_{\text{tgt}}$ successfully incorporates the changes dictated by $I$, and strictly evaluates the factual accuracy of any integrated content derived from the source paper $M$.
\textbf{(ii) Modification Scope.} This metric evaluates 
% the system's precision in localizing edits while preserving the integrity of the source poster. It verifies 
whether there are unnecessary or unintended modifications (e.g., hallucinations or collateral damage) in regions irrelevant to the instruction $I$.
\textbf{(iii) Visual Consistency \& Harmony.}
% This metric is based on instruction fulfillment and modification scope metrics, measures the aesthetic coherence of the modifications within the context of thec original poster. 
This metric\footnote{This metric is based on instruction fulfillment and modification scope metrics. High scores in instruction fulfillment and modification scope are prerequisites for a high score in this metric.} evaluates whether the newly integrated or modified elements align logically with the design (e.g., layout, typography, and style) of $P_{\text{src}}$. This dimension is particularly critical for handling abstract or underspecified instructions (e.g., \emph{``Add method section to top-right of the poster''}), where the system has to infer fine-grained visual details to maintain a professional and visually unified appearance
\section{Experiments}
% Due to the space limitation, we move additional experimental results to Appendix xxx, including xxx.
Due to space limitations, we move detailed case study to Appendix~\ref{sec:case_study_details}.
\subsection{Experimental Setup}
%\paragraph{Baselines.} We compare our method with several representative baselines that span different poster editing paradigms: \textbf{(i) Full XML Generation}, which directly regenerates the affected Office Open XML (OOXML) components from multi-modal inputs and performs edits at the file-structure level; \textbf{(ii) Python-PPTX Agent}, which generates and executes a monolithic python-pptx script to manipulate slide objects based on structured and visual inputs; \textbf{(iii) Direct Image Generation}, which formulates the task as image synthesis and produces a high-resolution PNG using a image generation and editing model without relying on intermediate presentation software or layout engines; \textbf{(iv) Talk-to-Your-Slides}~\cite{}, which edits slides through a structured JSON representation and a planner–executor architecture; and \textbf{(v) PPTC}~\cite{}, which translates natural language instructions into executable API sequences with support for both single-turn and multi-turn edits.with support for multi-turn reasoning.\scx{revise}

\paragraph{Baselines.} We compare our method with representative baselines from two categories. \textbf{(i) Regeneration-based methods} reformulate poster editing as regeneration, 
including \textit{XML Generation}
% , which regenerates Office Open XML (OOXML) components and re-packages them into slides,
and \textit{Direct Image Generation}.
% , which regenerate the poster image using image generation models for each edit. 
\textbf{(ii) Generic slide editing methods} perform slide editing on poster, including \textit{Direct Script-based Editing}
% , which directly generates executable \texttt{python-pptx} scripts for editing,
and agent-based approaches 
% such as 
\textit{Talk-to-Your-Slides}~\cite{jung2025talk} and \textit{PPTC}~\cite{guo2024pptc}. More details are provided in the Appendix~\ref{sec:appendix:baseline_detail}.
% We compare our method with several representative baselines that span different poster editing paradigms: \textbf{(i) XML Generation}, which regenerates the affected Office Open XML (OOXML) components and re-package the generated xml file into the slide; \textbf{(ii) Python-PPTX Agent}, which generates a monolithic Python script using the \texttt{python-pptx} library and executes it in sandbox environment to produce the final slide; \textbf{(iii) Direct Image Generation}, which formulates the task as a image regeneration problem, leveraging a Image Generation and Editing model to produce a flattened, high-fidelity image of the updated poster; \textbf{(iv) Talk-to-Your-Slides}~\cite{jung2025talk}, which utilizes a hierarchical architecture consisting of a high-level planner for instruction decomposition and a low-level executor that generates Python code to directly manipulate slide objects; and \textbf{(v) PPTC}~\cite{guo2024pptc}, which translates natural language instructions into executable API sequences through a logic-driven pipeline to create and edit PPT files. More details for the baseline are given in the Appendix.

\paragraph{Implementation Details.} Unless otherwise specified, all methods in our experiments are built upon advanced base models from the Gemini family~\cite{team2023gemini}. More implementation details are provided in Appendix~\ref{sec:appendix-impl}.
% 增加具体实现用的框架

\subsection{Main Results}

\begin{figure*}[!t]
    \centering
    \includegraphics[width=0.95\linewidth]{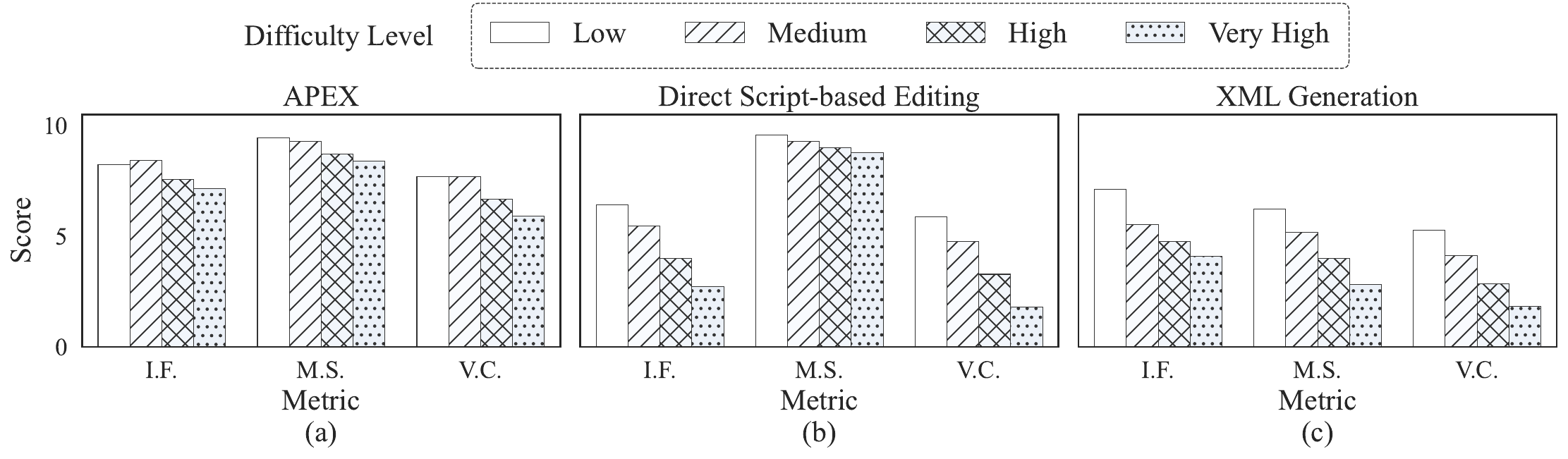}
    \caption{Performance across varying difficulty levels on three evaluation metrics. \textbf{I.F.}: Instruction Fulfillment, \textbf{M.S.}: Modification Scope, \textbf{V.C.}:
Visual Consistency.}
    \label{fig:difficulty}
\end{figure*}

\begin{table*}[!ht]
    \centering
    \scalebox{0.9}{
    \begin{tabular}{lcccc}
    \hline
    \textbf{Method} & \textbf{I.F. ($\uparrow$}) & \textbf{M.S. ($\uparrow$)} & \textbf{V.C. ($\uparrow$)} & \textbf{Cost (\$)} \\ 
    \hline
    \textbf{\method (Full)} & \textbf{7.97} & \underline{9.04} & \textbf{7.18} & 0.022 \\ 
    w/o Review           & \underline{7.32} & 8.58           & \underline{6.13} & 0.015 \\ 
    w/o Review \& APIs    & 5.12           & \textbf{9.14}  & 4.51           & 0.017 \\ 
    \hline
    \end{tabular}
    }
    \caption{Ablation study on the review-and-adjustment module and operation-to-API mapping.}
    \label{tab:ablation-coredesign}
\end{table*}

\paragraph{Overall Performance.} Table~\ref{tab:main_results} reports the performance of different methods,
% on instruction fulfillment, modification scope, and visual consistency, 
along with their execution error rate and monetary cost.
% Our method \method outperforms most competing approaches across all metrics and achieves the best results on instruction fulfillment and visual consistency, with a margin of approximately 15 percentage points. 
\method achieves the best performance on instruction fulfillment and visual consistency, with margins of approximately 15 percentage points. It also demonstrates comparable performance on modification scope.
% with the best performer.
The key insights are as follows. \textbf{(i)} Regeneration-based approaches, including direct image generation and XML generation, obtain significantly lower scores on modification scope, indicating their tendency to introduce unintended edits and hallucinations. \textbf{(ii)} PPTC and Talk-to-Your-Slides, despite being advanced slide editing agents, perform substantially worse in this setting, mainly due to the lack of explicit paper content extraction modules as well as limited capability in handling high-density information
% , complex visual elements, 
and compact layouts. 
% required by academic poster editing tasks. 
\textbf{(iii)} Direct script-based editing achieves the highest modification scope score but show much lower performance on instruction fulfillment and visual consistency, suggesting that many intended edits fail to be executed, as reflected by a high code execution error rate. \textbf{(iv)} \method demonstrates a favorable trade-off between performance and cost.

% \subsection{Fine-grained Analysis}

\paragraph{Difficulty-wise Performance.} Figure~\ref{fig:difficulty} illustrates the performance of different methods across editing instructions of varying difficulty levels. Overall, \method achieves superior performance compared to baseline methods across all difficulty levels. In particular, on high-difficulty samples, \method outperforms prior methods by a larger margin. In addition, the manually annotated difficulty levels align well with the observed performance trends, as scores
% on all metrics 
tend to decrease with increasing task difficulty. 
% This observation suggests that the manual difficulty annotations effectively reflect the underlying task complexity. 
Due to space limitations, additional results 
% and analyses 
are provided in Appendix~\ref{sec:appendix-difficulty}.

\subsection{Ablation Study}
% \begin{table}[!ht]
%     \centering
%     \begin{tabular}{|l|l|l|l|l|}
%     \hline
%         method & I.F. & M.S. & V.C. & cost\$ \\ \hline
%         ours & 7.97 & 9.04 & 7.18 & 0.0218435 \\ \hline
%         w/o reivew & 7.32 & 8.58 & 6.13 & 0.015006 \\ \hline
%         w/o review and api list & 5.12 & 9.14 & 4.51 & 0.0165975 \\ \hline
%     \end{tabular}
% \end{table}
% \begin{table}[!ht]
%     \centering
%     \scalebox{0.88}{
%     \begin{tabular}{lcccc}
%     \hline
%     \textbf{Method} & \textbf{I.F.} & \textbf{M.S.} & \textbf{V.C.} & \textbf{Cost (\$)} \\ 
%     \hline
%     \textbf{Ours (Full)} & \textbf{7.97} & 9.04 & \textbf{7.18} & 0.022 \\ 
%     w/o Review & 7.32 & 8.58 & 6.13 & 0.015 \\ 
%     w/o Review \& APIs & 5.12 & \textbf{9.14} & 4.51 & 0.017 \\ 
%     \hline
%     \end{tabular}
%     }
%     \caption{Ablation study of system components. I.F.: Instruction Fulfillment, M.S.: Modification Scope, V.C.: Visual Consistency.}
%     \label{tab:ablation-coredesign}
% \end{table}

% \textbf{Analysis of Core Design.} 
Table~\ref{tab:ablation-coredesign} presents the ablation results on two core design components: the review-and-adjustment module and the multi-level API-based editing mechanism. Three settings are considered: \textbf{(i)} the full model, \textbf{(ii)} a variant without the review-and-adjustment module, and \textbf{(iii)} a further ablated variant that additionally removes the operation-to-API sequence mapping and instead directly outputs \texttt{python-pptx} code for poster editing. Removing the review-and-adjustment module leads to a clear performance degradation, with the most substantial drop observed in visual consistency, indicating that incorporating visual feedback helps
the model 
% better assess intermediate outputs and 
improves both instruction adherence and layout-level adjustments. Further removing the operation-to-API sequence mapping results in pronounced declines in instruction fulfillment and visual consistency, while leading to increased execution cost, underscoring that the multi-level API-based editing mechanism is both more cost-efficient and more effective in enforcing instruction following. In this setting, the modification scope score slightly increases, which can be attributed to the reduced editing capability: fewer modifications are applied, making unintended changes less likely to occur. Overall, the ablation results demonstrate that
the two core designs
% the review-and-adjustment module and the multi-level API-based editing mechanism 
play complementary roles in improving instruction-following accuracy and visual quality. Due to the space limitation, we move the ablation results of base model to Appendix~\ref{sec:appendix-basemodel}. 

\section{Conclusion}
In this paper, we presented \method, the first agentic framework dedicated to interactive academic poster editing. Our framework consists of
% combined with 
two core designs, namely, 
a robust multi-level API suite and a review-and-adjustment mechanism, 
excels at manipulating high-density layouts of academic posters while ensuring localized, precise control. 
To support systematic research in this domain, we further introduced \bench, a 
% systematic 
benchmark of 514 editing instructions, and established a multi-dimensional VLM-as-a-judge evaluation protocol. 
Experimental results show that \method significantly outperforms both regeneration-based and generic slide-editing baselines in instruction fulfillment and visual consistency, while preserving 
% high score 
modification scope. 
Our work provides a reliable foundation for bridging the gap between automated drafting and professional-grade research dissemination.

\section*{Limitations}

Despite the promising results, our work has certain limitations that suggest directions for future research. 

First, due to constraints on computational resources, we did not explore more powerful frontier models (e.g., Gemini-3-Pro) for the agentic framework. Future work could conduct a more comprehensive exploration using these models to further enhance performance of the editing process. 

Second, our current framework lacks a mechanism to handle instructions that require external visual assets not present in the provided paper or poster. Extending the agent's capability to retrieve and integrate external multi-modal resources remains an important open challenge for academic poster editing task.

\section*{Ethical considerations}

We take ethical considerations seriously in the development of \method and \bench. 

\paragraph{Data Sourcing and Intellectual Property.} 

All research papers used in \bench were collected from top-tier AI conferences, including ICLR, ICML, and NeurIPS. These venues publish their proceedings under the Creative Commons Attribution 4.0 International License (CC BY 4.0).\footnote{\url{https://creativecommons.org/licenses/by/4.0/}} This license permits users to freely share (copy and redistribute) and adapt (remix, transform, and build upon) the material for any purpose, including scholarly and commercial use, provided that appropriate credit is given to the original authors. We strictly follow these terms by ensuring proper attribution in our dataset. The initial posters generated and the subsequent editing instructions are intended solely for research purposes to advance the field of academic poster editing. 

\paragraph{Annotator Recruitment and Ethics.} To ensure the high quality and reliability of the benchmark, we establish a rigorous annotation and verification protocol. We recruit a team of five master's and Ph.D. students as annotators, all proficient in English and possessing relevant expertise in academic publishing and design tools (e.g., PowerPoint). 

They are recruited as Research Assistants (RAs) and remunerated in accordance with the wage standards prescribed by local laws.

Before the annotation process, all annotators are informed about the task objectives and are required to strictly adhere to the annotation guidelines. 

\paragraph{Potential for Misuse.} While our system is designed to assist researchers in the labor-intensive process of poster creation, we acknowledge the potential risk of using such tools to generate misleading academic materials. We advocate for the responsible use of AI in scientific communication and emphasize that \method should be used as a collaborative assistant under human supervision rather than a tool for fully autonomous, unverified content generation.

% \section*{Acknowledgments}

% This document has been adapted
% by Steven Bethard, Ryan Cotterell and Rui Yan
% from the instructions for earlier ACL and NAACL proceedings, including those for
% ACL 2019 by Douwe Kiela and Ivan Vuli\'{c},
% NAACL 2019 by Stephanie Lukin and Alla Roskovskaya,
% ACL 2018 by Shay Cohen, Kevin Gimpel, and Wei Lu,
% NAACL 2018 by Margaret Mitchell and Stephanie Lukin,
% Bib\TeX{} suggestions for (NA)ACL 2017/2018 from Jason Eisner,
% ACL 2017 by Dan Gildea and Min-Yen Kan,
% NAACL 2017 by Margaret Mitchell,
% ACL 2012 by Maggie Li and Michael White,
% ACL 2010 by Jing-Shin Chang and Philipp Koehn,
% ACL 2008 by Johanna D. Moore, Simone Teufel, James Allan, and Sadaoki Furui,
% ACL 2005 by Hwee Tou Ng and Kemal Oflazer,
% ACL 2002 by Eugene Charniak and Dekang Lin,
% and earlier ACL and EACL formats written by several people, including
% John Chen, Henry S. Thompson and Donald Walker.
% Additional elements were taken from the formatting instructions of the \emph{International Joint Conference on Artificial Intelligence} and the \emph{Conference on Computer Vision and Pattern Recognition}.

\bibliography{main}

\appendix
%\begin{figure}
    %\centering
   % \includegraphics[width=0.5\linewidth]{category_bar_chart.pdf}
   % \caption{Enter Caption}
%\begin{figure}
        %\includegraphics[width=0.5\linewidth]{difficulty_donut_chart.pdf}
        %\caption{Enter Caption}
       % \label{fig:placeholder}
   % \end{figure}
       % \label{fig:placeholder}
%\end{figure}

\section{Benchmark Construction Details}
\label{sec:benchmark_details}
\subsection{Annotation Process}
\label{sec:annotation_process}
\paragraph{Annotator Recruitment and Ethics.} To ensure the high quality and reliability of the benchmark, we establish a rigorous annotation and verification protocol. We recruit a team of five master's and Ph.D. students as annotators, all proficient in English and possessing relevant expertise in academic publishing and design tools (e.g., PowerPoint). 
They are recruited as Research Assistants (RAs) and remunerated in accordance with the wage standards prescribed by local laws.
Before the annotation process, all annotators are informed about the task objectives and are required to strictly adhere to the annotation guidelines.

\paragraph{Task Decomposition.}
The annotators are tasked with two primary objectives: (i) \textit{Refining and Validating} the AI-generated instructions to ensure executability, and (ii) \textit{Labeling} the instructions according to our proposed taxonomy.

\paragraph{Detailed Annotation Guidelines.} 
To standardize the verification process, we provide annotators with explicit operational definitions for acceptance and rejection. The guidelines for the checklist are defined as follows:
\begin{itemize}
    \item \textbf{Validity Criteria:} Annotators verify the factual consistency between the instruction and the source paper. 
    An instruction is marked as \textit{Invalid} if it requests the generation of information, data points, or conclusions not explicitly supported by the text or figures in the source paper. This strict verification criterion guarantees that the dataset remains hallucination-free.

    \item \textbf{Layout Feasibility Protocol:} This criterion assesses the spatial viability of an instruction. Annotators are required to reject or rewrite instructions that necessitate placing elements in already fully occupied coordinates without a preceding ``clearing'' operation. When rewriting instructions involving spatial conflicts, annotators are instructed to decompose the target instruction into a logical sequence incorporating necessary prerequisite spatial adjustments to ensure non-overlapping placement.

    \item \textbf{Professional Rationality Standards:} This criterion serves as a filter for academic appropriateness. Annotators leverage their domain expertise to ensure that instructions are aligned with the norms of high-quality academic posters. Requests suggesting casual, overly decorative, or non-standard academic styles are systematically filtered out. 
\end{itemize}

\paragraph{Quality Assurance.} 
To ensure data quality, we adopt a rigorous collaborative review protocol. Each instruction is independently reviewed by at least two annotators. For the taxonomy labeling, we measure inter-annotator agreement and resolve discrepancies through discussion, ensuring that only high-quality and executable instructions are retained in the final benchmark.

\subsection{Detailed Taxonomy}
\label{sec:detailed_taxonomy}
\paragraph{Operation Category Details.}
Academic posters typically feature a distinct header and multiple logical sections containing dense text and figures.
To systematically characterize the diverse editing requirements of such information-dense artifacts, we establish a taxonomy of operation categories:

\begin{itemize}
\item \textbf{Text-Related:} This covers font attribute adjustments (e.g., size, color, bolding) and content-driven modifications, such as summarizing, rephrasing, or expanding specific text blocks, to highlight key scientific contributions.
\item \textbf{Image Adjustments:} This encompasses the interpretation of user editing intent to identify, extract, and integrate specific visual evidence or diagrams from the source paper.
\item \textbf{Shapes and Elements:} These components are essential for defining visual boundaries, grouping related content, and enhancing the overall aesthetic structure.
\item \textbf{Overall Layout:} Key operations include defining global whitespace, managing element hierarchy, and Section-level Management (e.g., ``\emph{swapping Method and Results}''), which requires a deep understanding of the academic narrative structure.
\end{itemize}

\paragraph{Difficulty Levels.} 
To systematically characterize task complexity, we classify instructions into four levels based on edit complexity:
  \begin{itemize}
      \item \textbf{Low:} Operations with explicit parameters and straightforward execution logic, requiring limited spatial reasoning or semantic interpretation.
      (e.g., ``\emph{increase font size of all bullet heading, which are in front of `: ', to make them more prominent.}'')
      \item \textbf{Medium:} Multi-step operations involving spatial adjustments, element coordination, or paper-dependent content extraction, with moderate reasoning complexity.
      (e.g., ``\emph{Create a new section summarizing the main contribution of the paper in left-bottom blank, aligning the content with the main argument presented in the paper.}'')
      \item \textbf{High:} Complex operations involving substantial layout reconfiguration or deep reasoning.
      (e.g., ``\emph{Consolidate the `Problem Context' and the `RNAInterAct Dataset' sections into a single `In a nutshell' section. Briefly summarize the key problems and introduce the dataset. Then add a dark blue background bar under this section title and set appropriate text color to improve contrast. Add light color background bar to the other section headings.}'')
      \item \textbf{Very High:} Composite operations requiring the orchestration of global planning and fine-grained manipulation, guided by long-horizon reasoning.
      (e.g., ``\emph{Restructure the layout from a 3-column format to a 4-section modular layout (Introduction → Architecture → Experiments → Results/Summary) to improve narrative flow and navigation.}'')
  \end{itemize}

\paragraph{Abstraction Level.} 
To systematically evaluate the agent's adaptability to command ambiguity, we categorize instructions into two distinct levels of abstraction:
\begin{itemize}
    \item \textbf{Concrete Instructions:} These instructions impose explicit constraints with quantitative parameters, leading to a uniquely determined editing outcome. 
    (e.g., ``\emph{Change the font size of the title to 48pt and move it to the top-center.}'')
    \item \textbf{Abstract Instructions:} These instructions reflect subjective user intents or high-level aesthetic goals, admitting multiple valid solutions and requiring the agent to infer specific visual operations. 
    (e.g., ``\emph{Adjust the layout to reduce whitespace.}'')
\end{itemize}

\paragraph{Paper Dependency.}
To distinguish between purely visual edits and content-aware edits, we label instructions based on their reliance on the source paper:
\begin{itemize}
    \item \textbf{Paper-Related:} Instructions that mandate cross-modal reasoning to align editing decisions with the source paper's semantic context and factual details.
    \item \textbf{Paper-Independent:} Instructions that can be resolved solely using the visual information present on the poster canvas.
\end{itemize}

Table~\ref{tab:stats} summarizes the distribution of instructions across all four taxonomic dimensions, highlighting the dataset's diversity.

% 三线表
\begin{table}[!t]
\centering
\small
\begin{tabular}{llr}
\toprule
\textbf{Dimension} & \textbf{Subset} & \textbf{Ratio (\%)} \\
\midrule
Operation Category & Text-related & 79.77 \\
 & Overall layout & 59.34 \\
 & Image adjustments & 47.67 \\
 & Shapes and elements & 33.27 \\
\midrule
Difficulty & Medium & 36.19 \\
 & High & 29.77 \\
 & Low & 21.98 \\
 & Very High & 12.06 \\
\midrule
Abstraction & Concrete & 71.60 \\
 & Abstract & 28.40 \\
\midrule
Dependency & Paper-related & 57.20 \\
 & Paper-independent & 42.80 \\
\bottomrule
\end{tabular}
\caption{Detailed statistics of \bench. The dataset contains a total of 514 instructions. Operation categories are not mutually exclusive.}
\label{tab:stats}
\end{table}

\begin{table*}[htbp]
\centering

\label{tab:benchmark_comparison}
% \small
\resizebox{\textwidth}{!}{
\begin{tabular}{lp{6cm}p{7.5cm}}
\toprule
\textbf{Benchmark} & \textbf{Scenario Focus} & \textbf{Evaluation Coverage and Gaps} \\ \midrule

PPTC~\cite{guo2024pptc} & 
Multi-turn instruction completion for PowerPoint (PPT) creation and editing. & 
Evaluates functional API accuracy via the PPTX-Match system. \textbf{Gap}: Overlooks design aesthetics and visual logic. \\ \addlinespace

PPTC-R~\cite{zhang2024pptc} & 
Robustness of PPT task completion against adversarial instructions and software versioning. & 
Covers sentence, semantic, and multi-language robustness. \textbf{Gap}: Performance significantly degrades in low-resource languages. \\ \addlinespace

TSBench~\cite{jung2025talk} & 
Efficient, low-cost slide editing by leveraging structured application objects. & 
Focuses on instruction fidelity, processing latency, and operational cost. \textbf{Gap}: Primarily targets localized edits; lacks macro-structural planning. \\ \addlinespace

Paper2Poster~\cite{pang2025paperposter} & 
End-to-end automated generation of academic posters from scientific papers (extreme compression). & 
Visual quality (CLIP), textual coherence, and ``Paper Quiz'' for knowledge transfer. \textbf{Gap}: Serial refinement limits generation efficiency. \\ \addlinespace

P2P~\cite{sun2025p2pautomatedpapertopostergeneration} & 
Automated generation of responsive, high-quality HTML-rendered academic posters. & 
Universal evaluation (e.g., whitespace balance) and fine-grained human-annotated checklists. \textbf{Gap}: Format limited to Web-based rendering. \\ \addlinespace

PPTBench~\cite{huang2025pptbenchholisticevaluationlarge} & 
Holistic evaluation of layout and design understanding for PPT tasks. & 
11 sub-tasks from detection to generation using dual-modality inputs (JSON \& Image). \textbf{Gap}: Highlights a major ``semantic-spatial gap'' in layout reasoning. \\ \addlinespace

PPTArena~\cite{2025pptarena} & 
Reliable in-place editing of real-world decks under natural-language instructions. & 
Over 800 edits across text, charts, tables, and styles using a dual VLM-as-a-judge for visual and semantic intent. \textbf{Gap}: General domain focus; limited deep context reasoning from scientific documents. \\ \addlinespace

\textbf{\bench (Ours)} & 
High-density academic posters using native \textbf{PPTX} objects with 514 human-verified instructions, which incorporating paper-related instruction for better refine the poster. &With VLM-as-a-judge to evaluate across three dimensions: instruction fulfillment, modification scope, and visual consistency \& harmony. \textbf{Strength}: Bridges the gap between automated drafts and professional author-level standards. \\ \bottomrule
\end{tabular}
}
\caption{Comparison of various benchmarks for automated presentation and poster generation/editing.}
\end{table*}

\begin{figure*}[!t]
    \centering
    \includegraphics[width=0.95\linewidth]{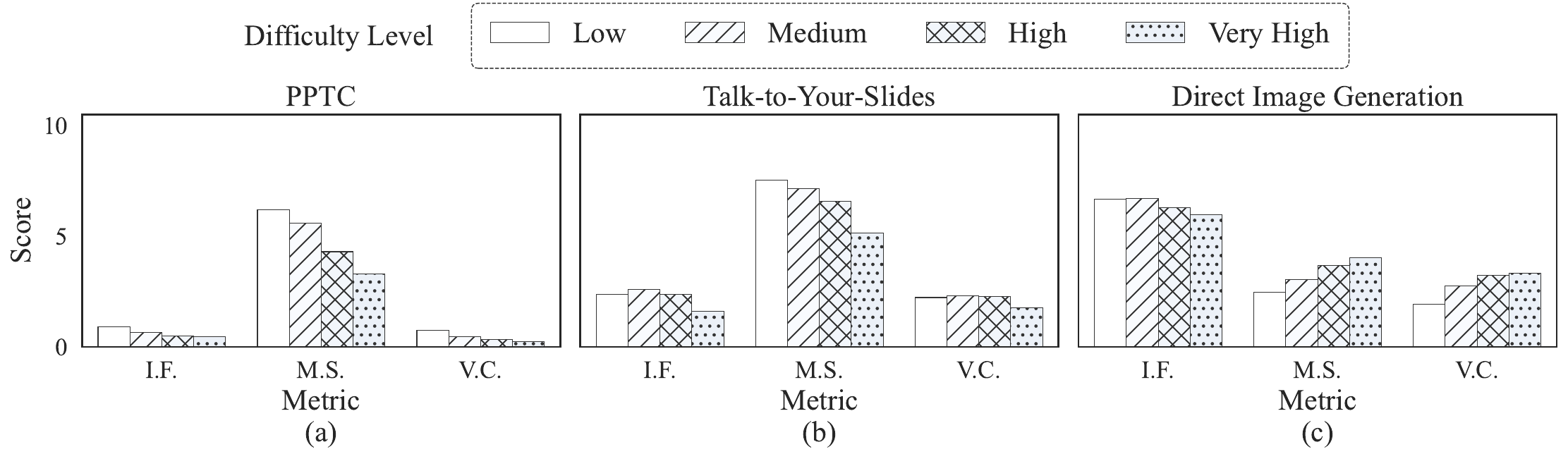}
    \caption{Performance across varying difficulty levels on three evaluation metrics.}
    \label{fig:difficulty-2}
\end{figure*}

\begin{table*}[!ht]
    \centering
    \scalebox{0.9}{
    \begin{tabular}{lcccc}
    \hline
    \textbf{Model} & \textbf{I.F. ($\uparrow$)} & \textbf{M.S. ($\uparrow$)} & \textbf{V.C. ($\uparrow$)} \\ 
    \hline
    \textbf{Gemini-3-Flash-Preview} & \textbf{7.97} & \textbf{9.04} & \textbf{7.18} \\ 
    Qwen3-VL-Plus & \underline{5.92} & \underline{7.27} & \underline{4.55} \\ 
    Qwen3-VL-30B-A3B & 4.44 & 5.99 & 3.05 \\ 
    \hline
    \end{tabular}
    }
    \caption{Performance of \method\ on different base models.}
    \label{tab:ablation-basemodel}
\end{table*}

\section{More Experiment Details}
\subsection{Detailed Baseline Description}
\label{sec:appendix:baseline_detail}
\begin{itemize}
    \item \textbf{XML Generation:} This approach manipulates the document's underlying OOXML structure. The model is provided with a multi-modal context, including user instructions, a JSON-formatted poster representation, visual states, and raw XML snippets. The task is to regenerate the entire Office Open XML (OOXML) payload for the affected components. By generating complete, syntactically valid XML blocks that reflect the requested edits, the agent performs structural modifications before re-packaging them into the slide.

    \item \textbf{Direct Script-based Editing:} This approach generates a single, monolithic Python script utilizing the \texttt{python-pptx} library. It incorporates multi-modal inputs, including JSON-formatted metadata and base64-encoded visual representations of the existing slides. The generated code is executed within a sandboxed environment to produce the final PPTX file and its corresponding preview image.

    \item \textbf{Direct Image Generation:} This method frames the task as a direct image synthesis problem. An image generation and editing model (e.g., Gemini-3-Pro-Image) receives user instructions, the original poster image, and relevant paper content as visual and textual context. The model then generates a flattened, high-fidelity PNG image of the updated poster in a single pass.

    \item \textbf{Talk-to-your-slides \cite{jung2025talk}:} This method is an agent-based framework designed for efficient slide editing. Departing from traditional GUI-based agents that rely on pixel-level interactions, it converts PowerPoint slides into a structured JSON representation. It utilizes a hierarchical architecture: a high-level planner for instruction decomposition and a low-level executor that generates Python code for object manipulation. This design ensures high precision in text, layout, and formatting while minimizing computational latency and costs.

    \item \textbf{PPTC \cite{guo2024pptc}:} PPTC employs a logic-driven pipeline to transform natural language instructions into executable API sequences. A key strength lies in its dual-mode processing: (1) \textit{Single-Turn Instruction Mapping}, which maps specific intents to discrete API sequences for isolated tasks; and (2) \textit{Multi-Turn Session Reasoning}, which maintains a ``session state'' for incremental edits. This allows the agent to interpret instructions within the context of previous operations
    , ensuring cumulative modifications remain consistent.
\end{itemize}

\subsection{More Implementation Details}
\label{sec:appendix-impl}
The \textit{Direct Image Generation} baseline uses Gemini-3-Pro-Image-Preview, while all other methods, as well as the judge model, are based on Gemini-3-Flash-Preview. The temperature of the judge model is set to 0 to ensure deterministic evaluations, whereas a temperature of 0.1 is used for all other models.

\subsection{Difficulty-wise Performance}
\label{sec:appendix-difficulty}

Figure~\ref{fig:difficulty-2} further analyzes the performance of three additional baselines across different difficulty levels. For PPTC and Talk-to-Your-Slides, performance generally decreases as instruction difficulty increases, indicating a negative correlation between task difficulty and editing quality. In contrast, \textit{Direct Image Generation} exhibits an unexpected decreasing trend on both modification scope and visual consistency \& harmony as difficulty increases. Further analysis suggests that this behavior stems from the model’s tendency to perform global modifications: as instruction difficulty rises, editing requests typically involve a larger number of poster elements, making it less likely for unnecessary changes to be introduced beyond the specified edits. As a result, fewer unintended modifications are made for higher-difficulty instructions, leading to lower modification scope scores. Since visual consistency \& harmony is derived from Modification Scope, it shows a similar downward trend.

\subsection{The Impact of Base Model Choice}
\label{sec:appendix-basemodel}

Table \ref{tab:ablation-basemodel} shows that Gemini-3-Flash-Preview consistently outperforms the other base models across all three metrics. In contrast, Qwen3-VL-Plus\cite{bai2025qwen3vltechnicalreport} and Qwen3-VL-30B-A3B exhibit notably lower performance, with the largest gaps observed in visual consistency, indicating a higher tendency to introduce visually incoherent edits. These results suggest that base model capability significantly influences editing quality, particularly for maintaining coherent layouts.

\section{Case Study}
\label{sec:case_study_details}
\subsection{Comparison Study}

% --- 正文描述部分 ---
Here is a case of editing a poster generated by PosterGen (Figure~\ref{fig:sub1}) according to a complex user's instruction\footnote{Detail of this user instruction: Swap the positions of the ``Semantics-Aware Method'' section and the ``Key Findings'' section. Then restructure the ``Key Findings'' section into two equally wide left and right columns: place the textual content of the ``Key Findings'' section in the left column and the images in the right column. Remove the images from the ``Theoretical Results'' section. Insert a new section named ``Semantic-Aware Robustness'' between the two sections in the middle column, using two separate lines of text to introduce ``Adversarial Example'' and ``Over-robust Example'' respectively. Finally, adjust the spacing of the middle column so that it is visually appropriate, with no overlap or overflow.} involving swapping sections, restructuring content into columns, removing specific images, and adjusting vertical spacing.
As shown in Figure~\ref{fig:main_four_images}, the \textbf{Direct Image Generation} approach (Figure~\ref{fig:sub2}) fails to restructure the ``Key Findings'' section into two columns as requested and leaves inappropriate gaps in the middle column. Furthermore, we observe that images generated via this method suffer from severe text distortion and visual artifacts, as show in Figure~\ref{fig:sub5} .
The \textbf{XML Generation} method (Figure~\ref{fig:sub3}) exhibits significant visual artifacts: it distorts the ICLR logo size, loses the affiliation logos, and fails to correctly swap the section positions or follow the layout constraints. 
In contrast, our method, \textbf{\method} (Figure~\ref{fig:sub4}), not only accurately satisfies all complex  requirements including section swapping, column restructuring, and content insertion, but also preserves the integrity of unrelated elements like logos and header information without any overlap or overflow.

% --- 图片排版部分 ---

\subsection{Case Study On Review-And-Adjustment Mechanism}

\begin{figure*}[!t] 
    \centering 
    \begin{subfigure}[b]{0.48\textwidth} 
        \centering
        \includegraphics[width=\linewidth]{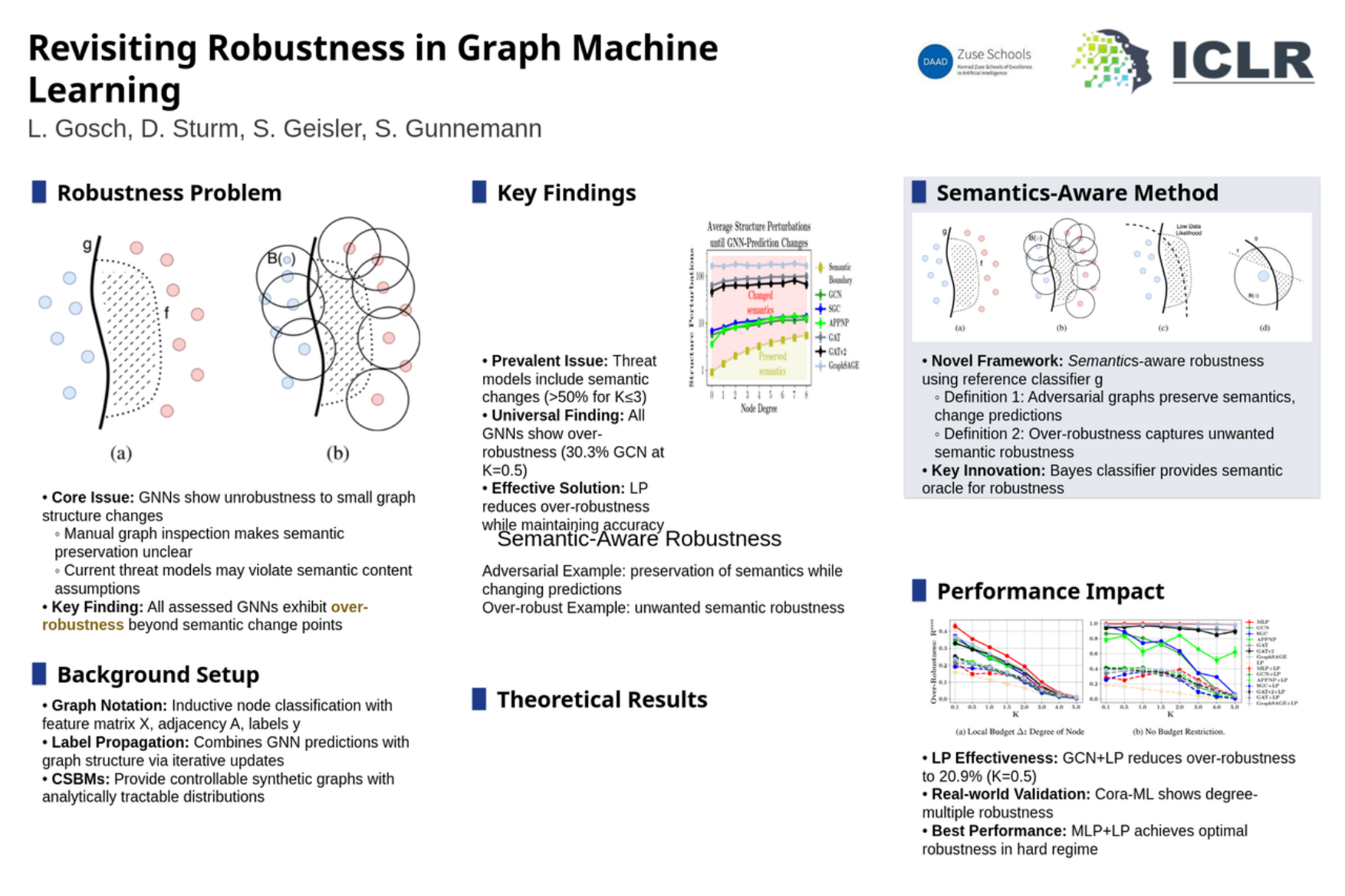} 
        \caption{Initial Edit}
        \label{fig:review_initial}
    \end{subfigure}
    \hfill 
    \begin{subfigure}[b]{0.48\textwidth}
        \centering
        \includegraphics[width=\linewidth]{1-4.pdf} 
        \caption{After Review-And-Adjustment}
        \label{fig:review_final}
    \end{subfigure}
    \caption{Effectiveness of the Review-And-Adjustment mechanism in adhering to complex instruction.}
\end{figure*}

The Review-And-Adjustment mechanism is critical for adhering complex instruction relation to intricate layout adjustment. As seen in the comparison between the initial edit result (Figure~\ref{fig:review_initial}) and the final result (Figure~\ref{fig:review_final}) under the same instruction in previous subsection, the adjustment phase corrects severe alignment issues and ensures the visual hierarchy remains professional.

\subsection{Significant Hallucinations In Regeneration-Based Methods}

\begin{figure*}[htbp] 
    \centering 

    % --- 第一行 ---
    \begin{subfigure}[b]{0.48\textwidth} 
        \centering
        \includegraphics[width=\linewidth]{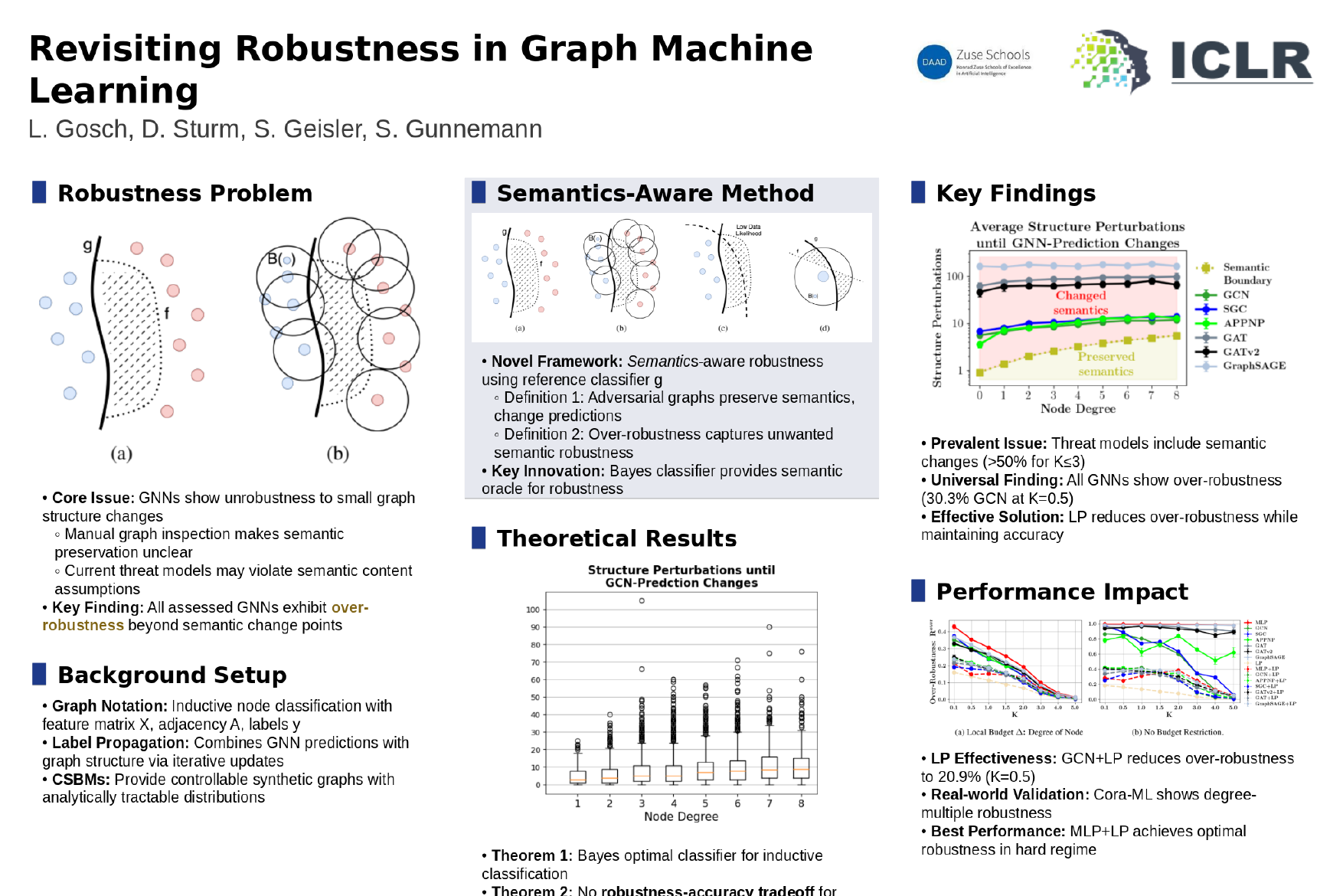} 
        \caption{Original Poster (PosterGen)}
        \label{fig:sub1}
    \end{subfigure}
    \hfill 
    \begin{subfigure}[b]{0.48\textwidth}
        \centering
        \includegraphics[width=\linewidth]{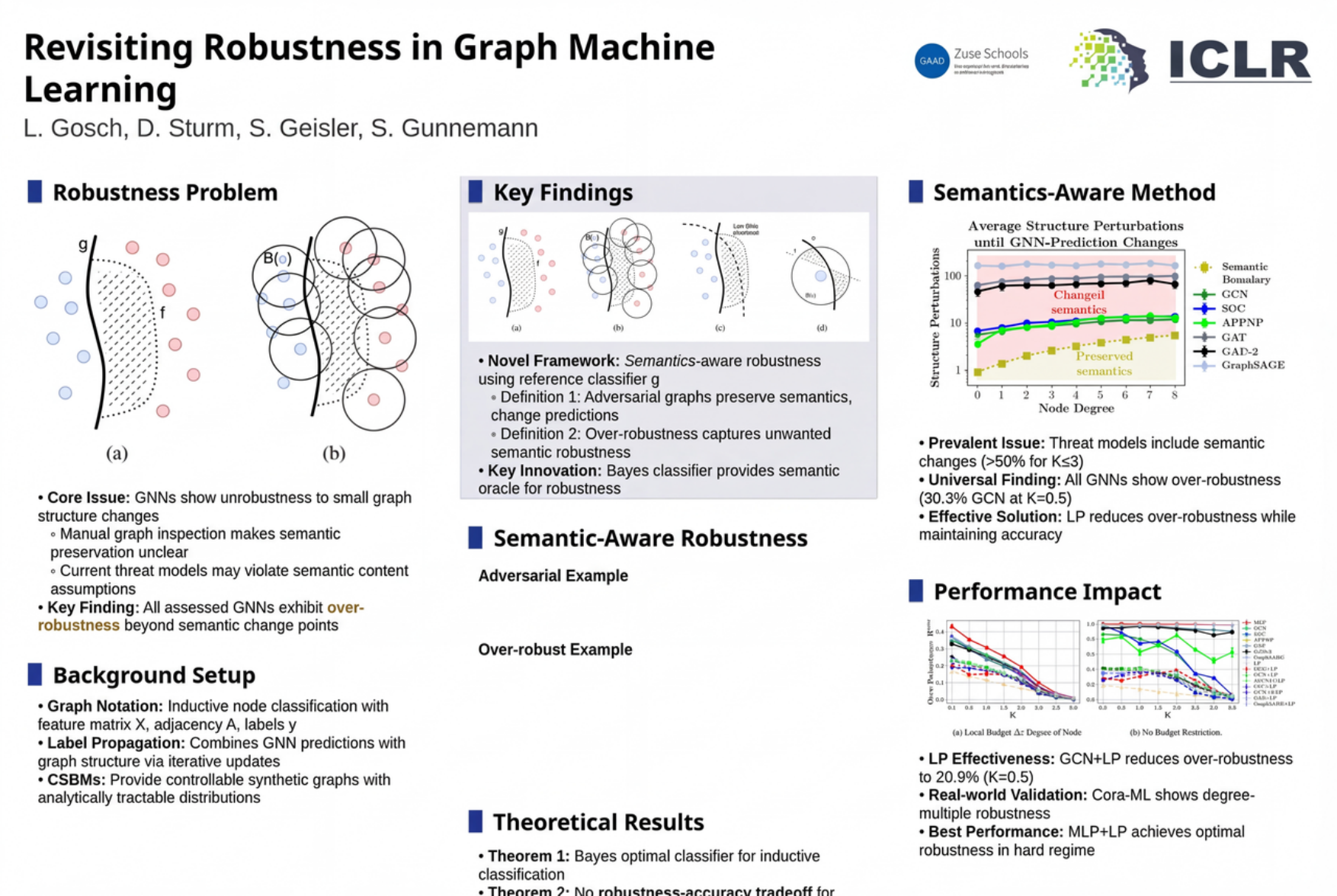} 
        \caption{Direct Image Generation}
        \label{fig:sub2}
    \end{subfigure}

    % --- 第二行 ---
    \begin{subfigure}[b]{0.48\textwidth}
        \centering
        \includegraphics[width=\linewidth]{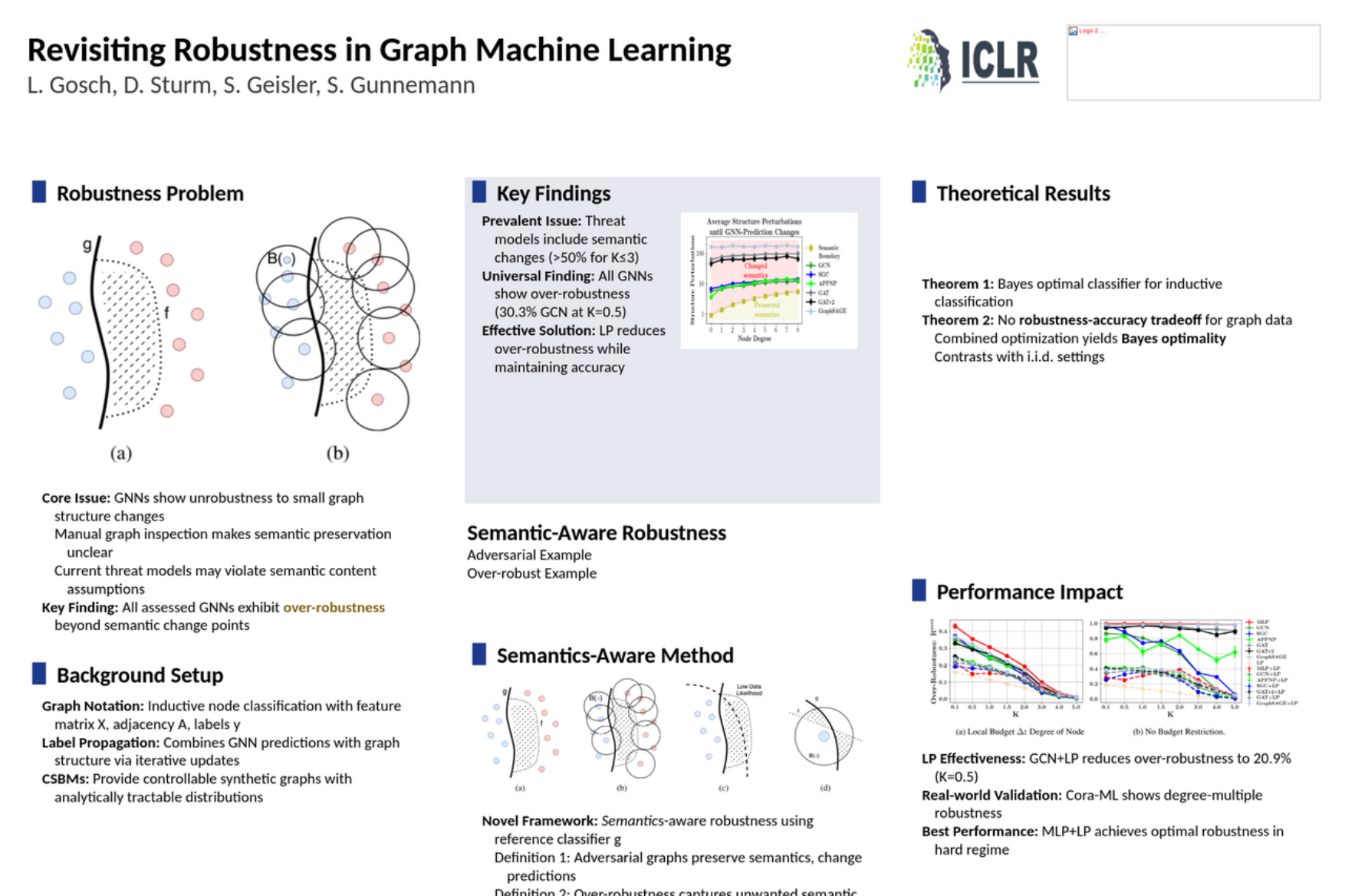} 
        \caption{XML Generation}
        \label{fig:sub3}
    \end{subfigure}
    \hfill
    \begin{subfigure}[b]{0.48\textwidth}
        \centering
        \includegraphics[width=\linewidth]{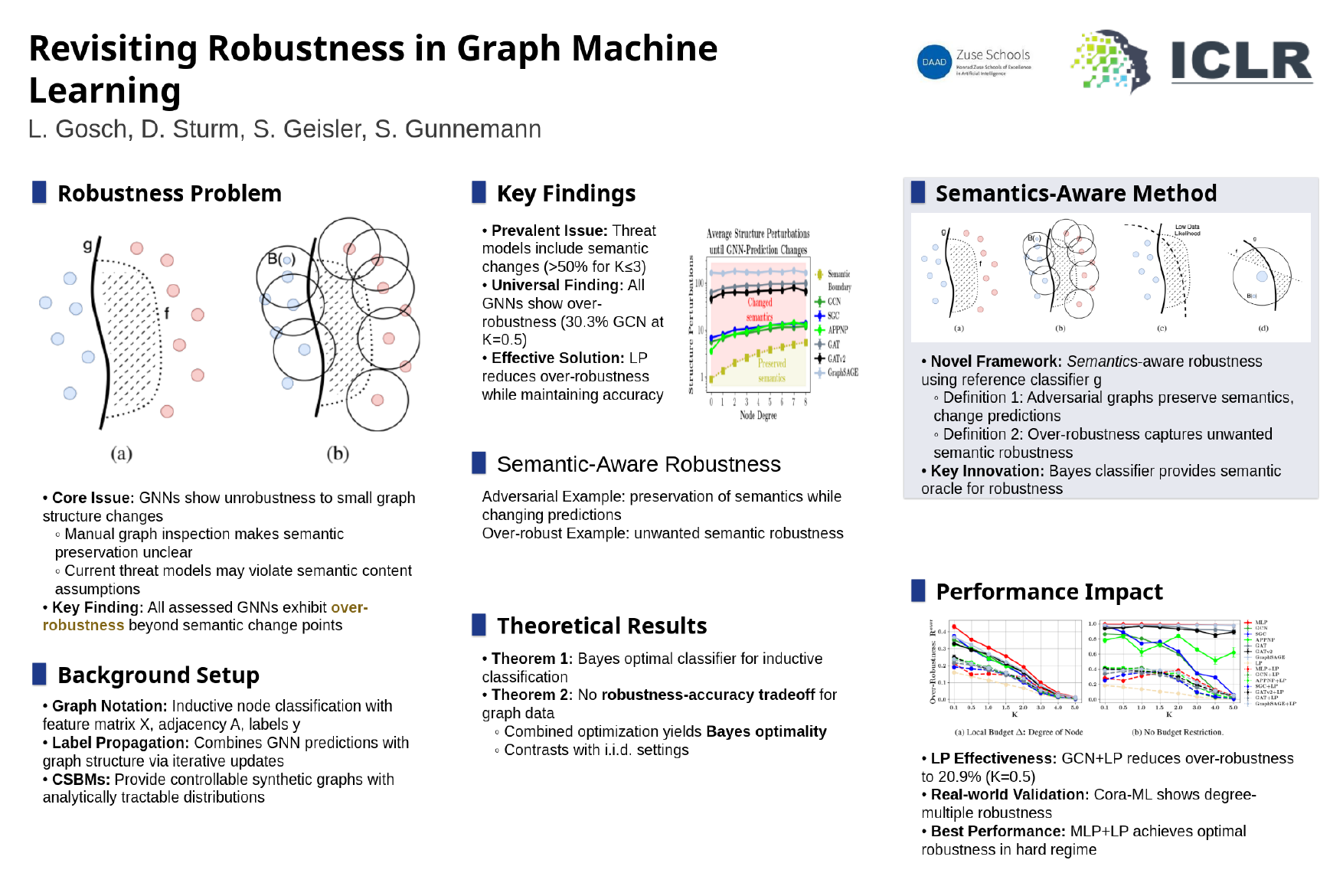} 
        \caption{\method (Ours)}
        \label{fig:sub4}
    \end{subfigure}
    % --- 第三行 ---
    \begin{subfigure}[b]{0.96\textwidth}
        \centering
        \includegraphics[width=\linewidth]{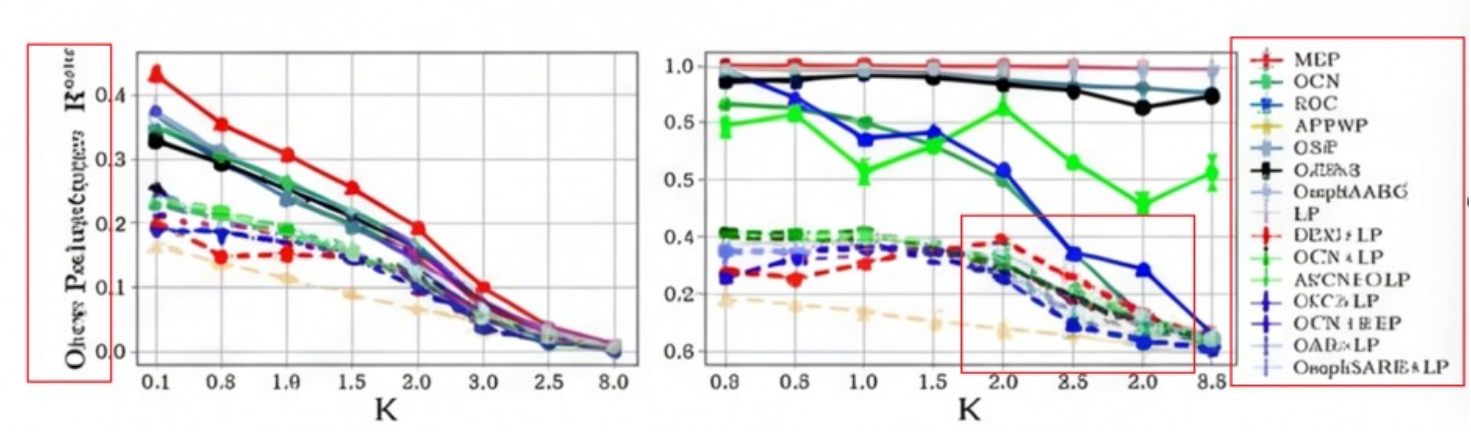} 
        \caption{Text Distortion and Visual Artifacts in Direct Image Generation}
        \label{fig:sub5}
    \end{subfigure}
    
    \caption{Comparison of different methods under a complex editing instruction, involving swapping sections, restructuring content into columns, removing specific images, and adjusting vertical spacing.} 
    \label{fig:main_four_images} 
\end{figure*}

\begin{figure*}[t] 
    \centering 
    \begin{subfigure}[b]{0.48\textwidth} 
        \centering
        \includegraphics[width=\linewidth]{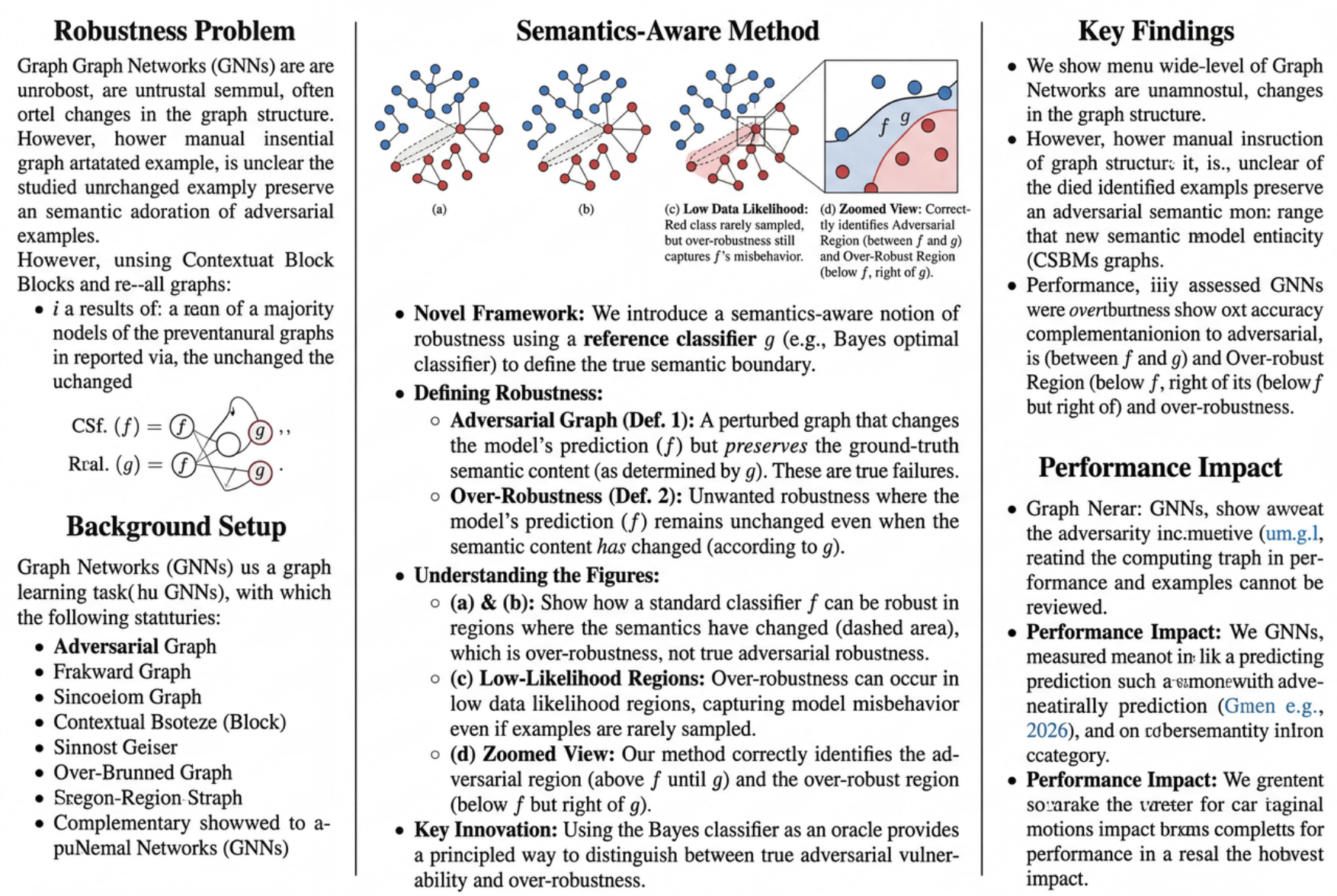} 
        \caption{Direct Image Generation}
        \label{fig:hallucination_initial}
    \end{subfigure}
    \hfill 
    \begin{subfigure}[b]{0.48\textwidth}
        \centering
        \includegraphics[width=\linewidth]{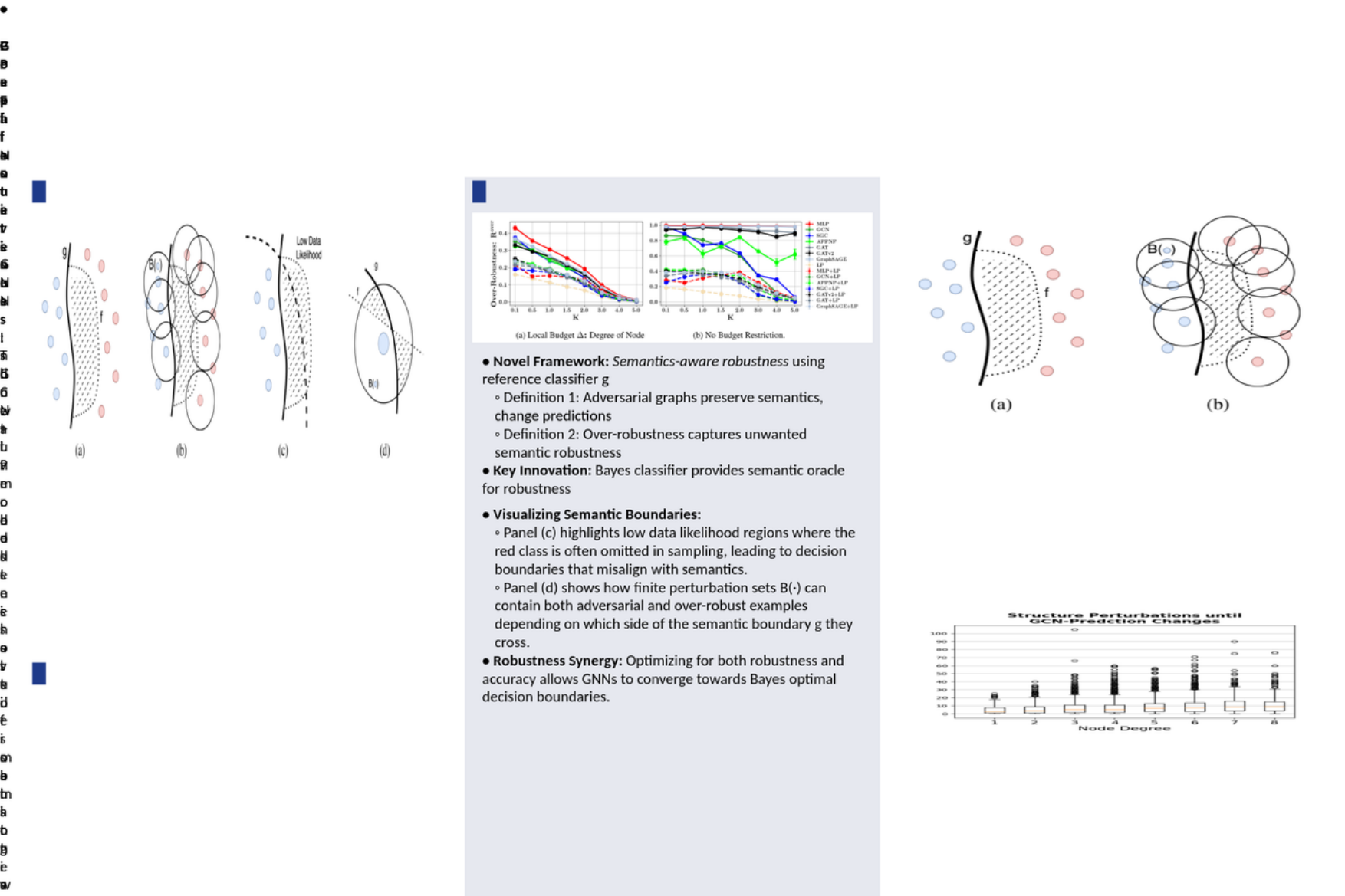} 
        \caption{XML Generation}
        \label{fig:hallucination_final}
    \end{subfigure}
    \caption{Severe hallucinations and style degradation in regeneration-based methods.}
\end{figure*}

Regeneration-based methods (Direct Image and XML) often suffer from significant hallucinations. Here are Direct Image Generation (Figure~\ref{fig:hallucination_initial}) and XML Generation (Figure~\ref{fig:hallucination_final}) results of editing the poster generated by PosterGen (Figure~\ref{fig:sub1}) according to user's instruction: ``Remove theoretical results section, expand semantics-aware method section to fill the middle column, specifically, illustrate the c and d figure in the middle column more detailed by adding appropriate text and then revise original text to reach a appropriate logical flow''. In this case, these methods fails to maintain the textual integrity and layout consistency of the original poster.

\section{Prompts}
\subsection{APEX Prompts}
\begin{figure*}[!t]
    \centering
    \includegraphics[width=1\linewidth]{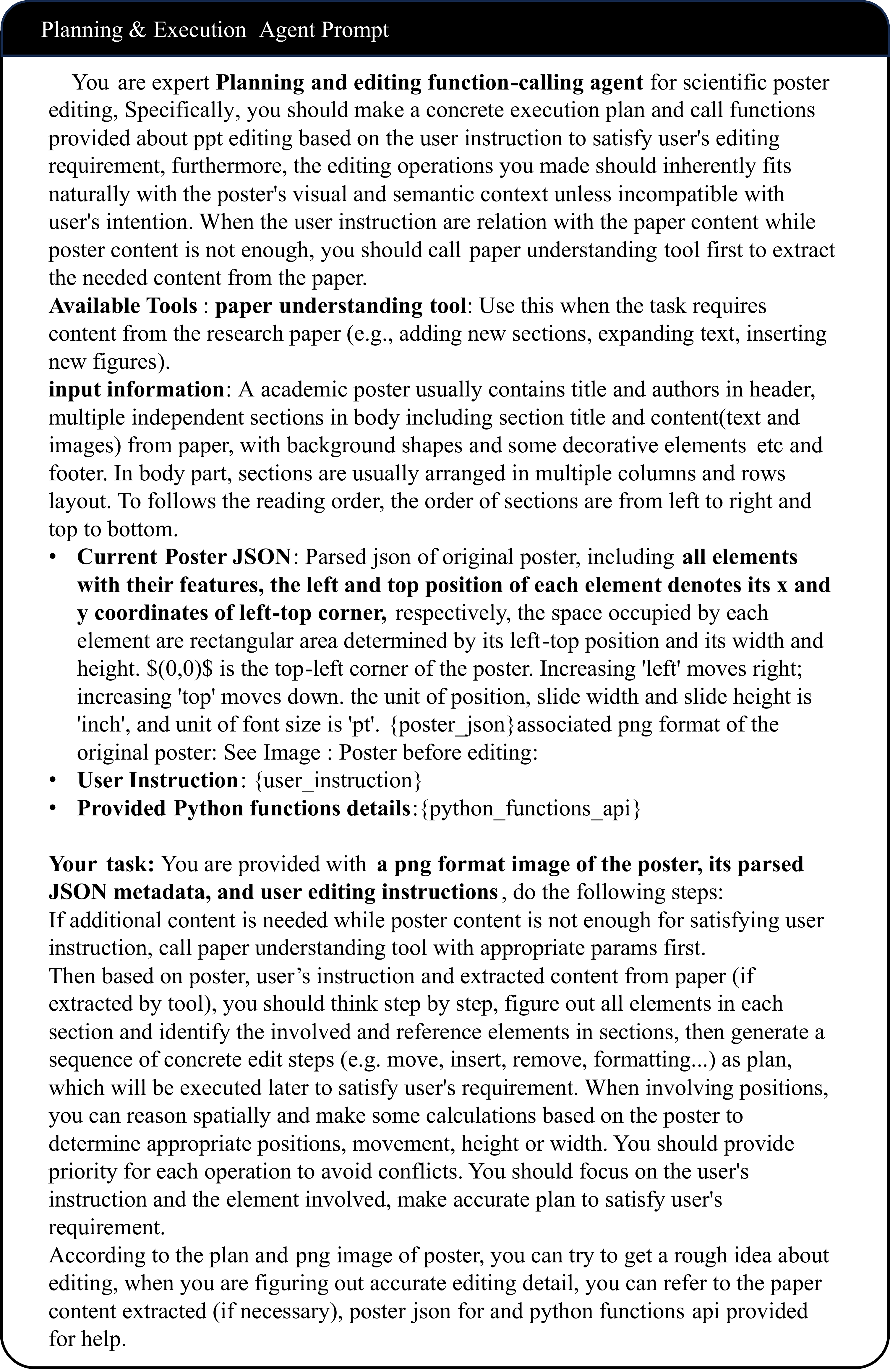}
    \caption{Prompt for Planning \& Execution Agent to plan and execute API.} 
    \label{fig:planner_prompt1}
\end{figure*}

\begin{figure*}[!t]
    \centering
    \includegraphics[width=1\linewidth]{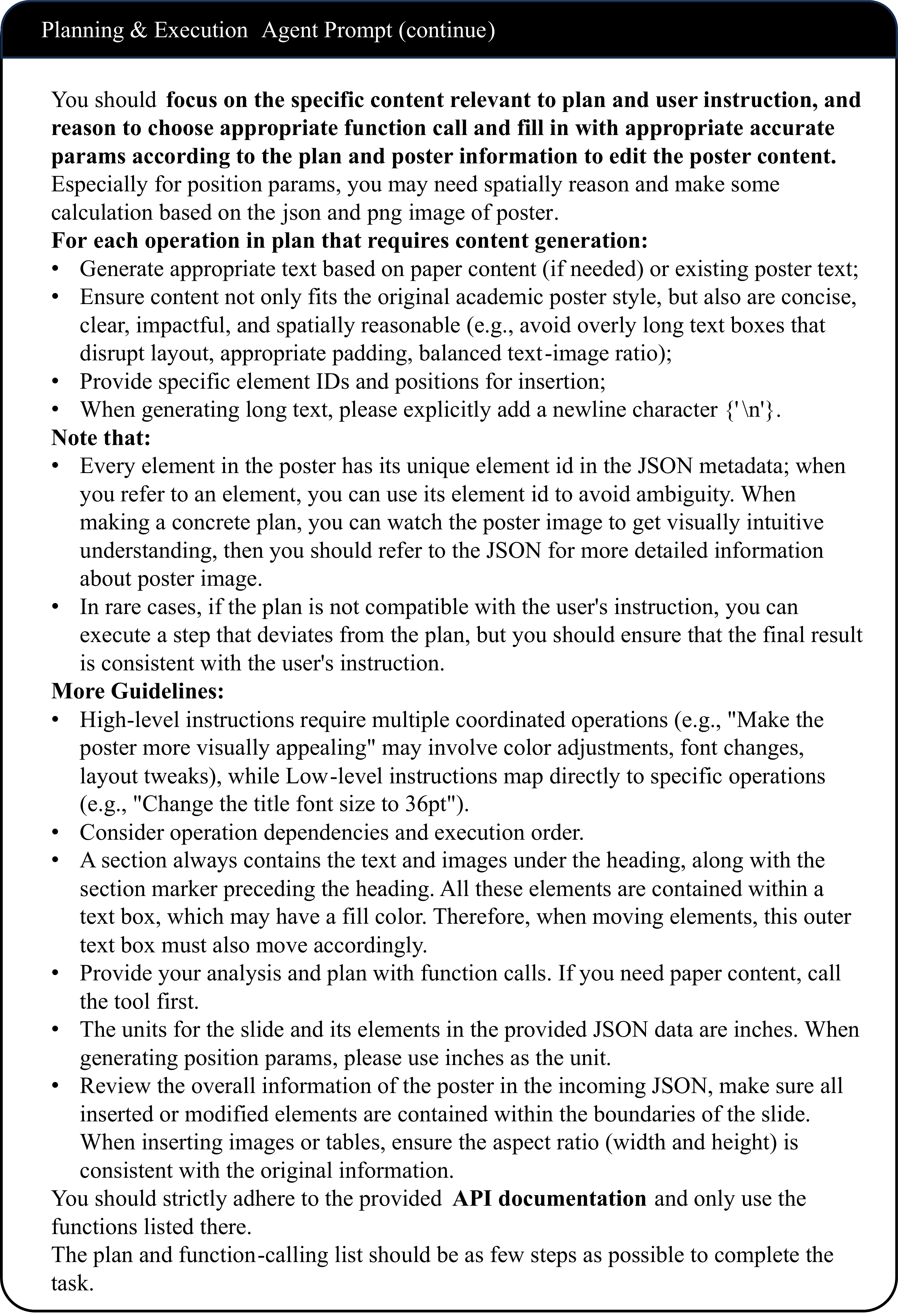}
    \caption{Prompt for Planning \& Execution Agent to plan and execute API (continue).} 
    \label{fig:planner_prompt2}
\end{figure*}

\begin{figure*}[!t]
    \centering
    \includegraphics[width=1\linewidth]{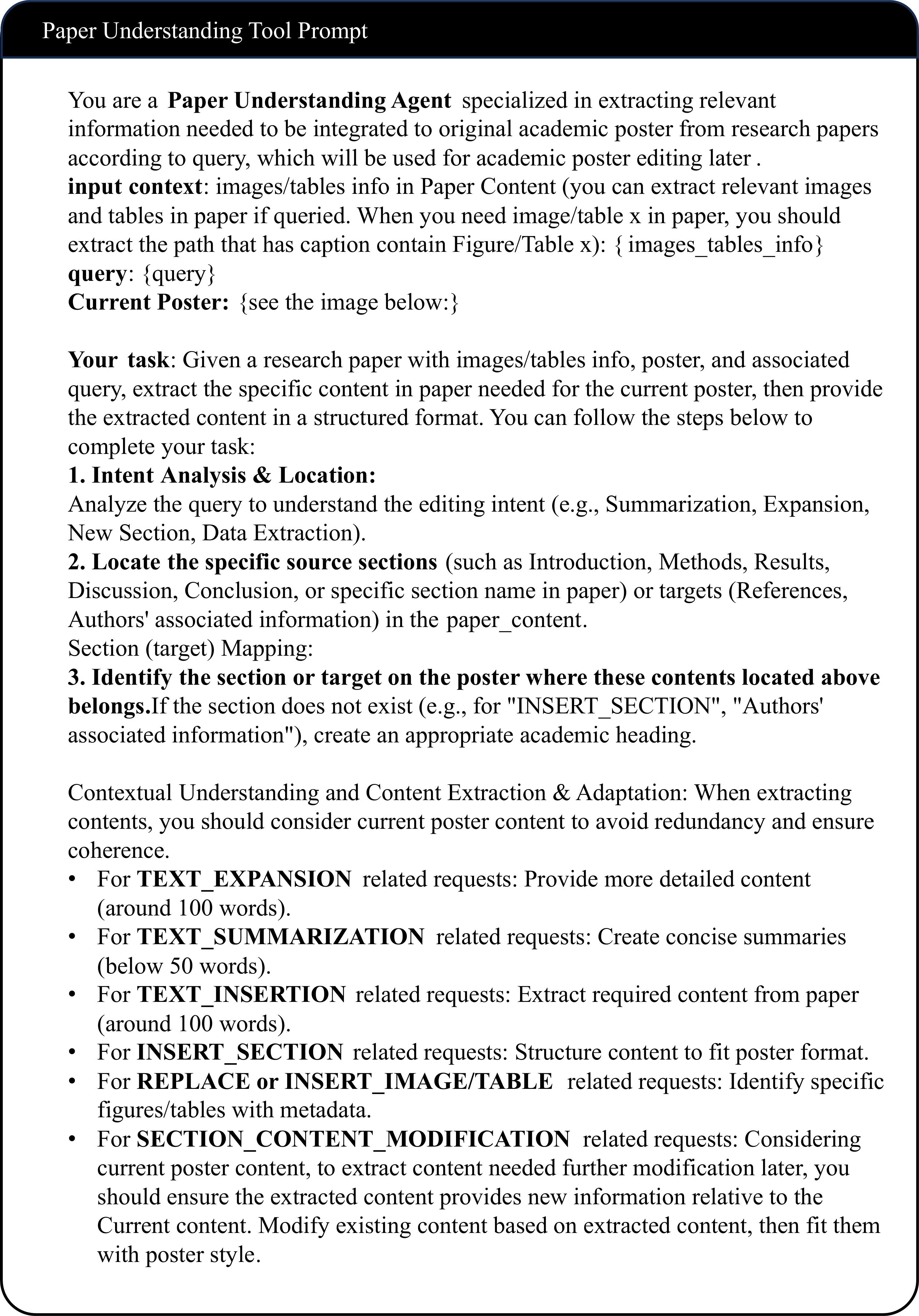}
    \caption{Prompt for paper understanding tool to extract paper-related content.} 
    \label{fig:planner_prompt3}
\end{figure*}

\begin{figure*}[!t]
    \centering
    \includegraphics[width=1\linewidth]{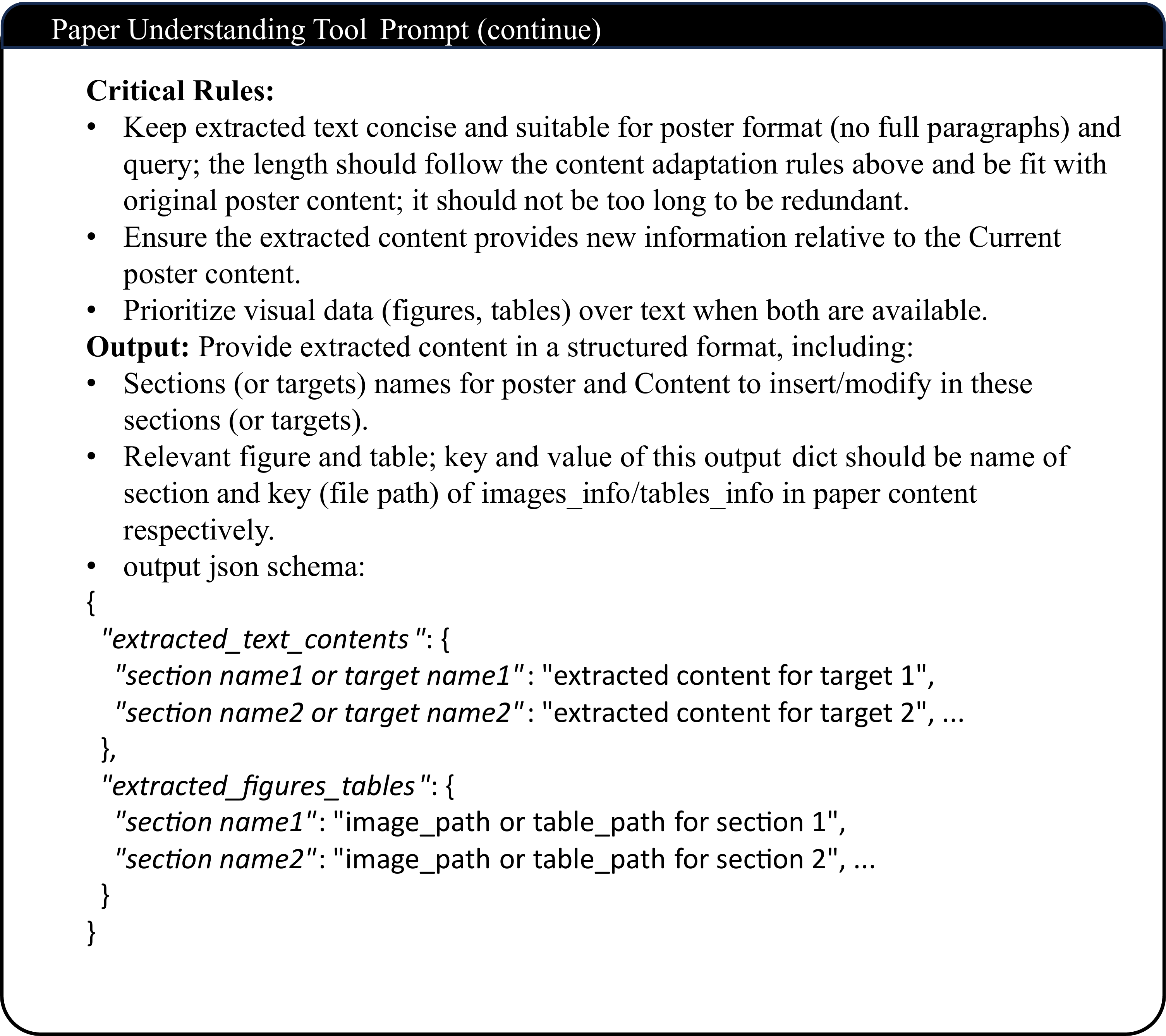}
    \caption{Prompt for paper understanding tool to extract paper-related content (continue).} 
    \label{fig:planner_prompt4}

\end{figure*}

\begin{figure*}[!t]
    \centering
    \includegraphics[width=1\linewidth]{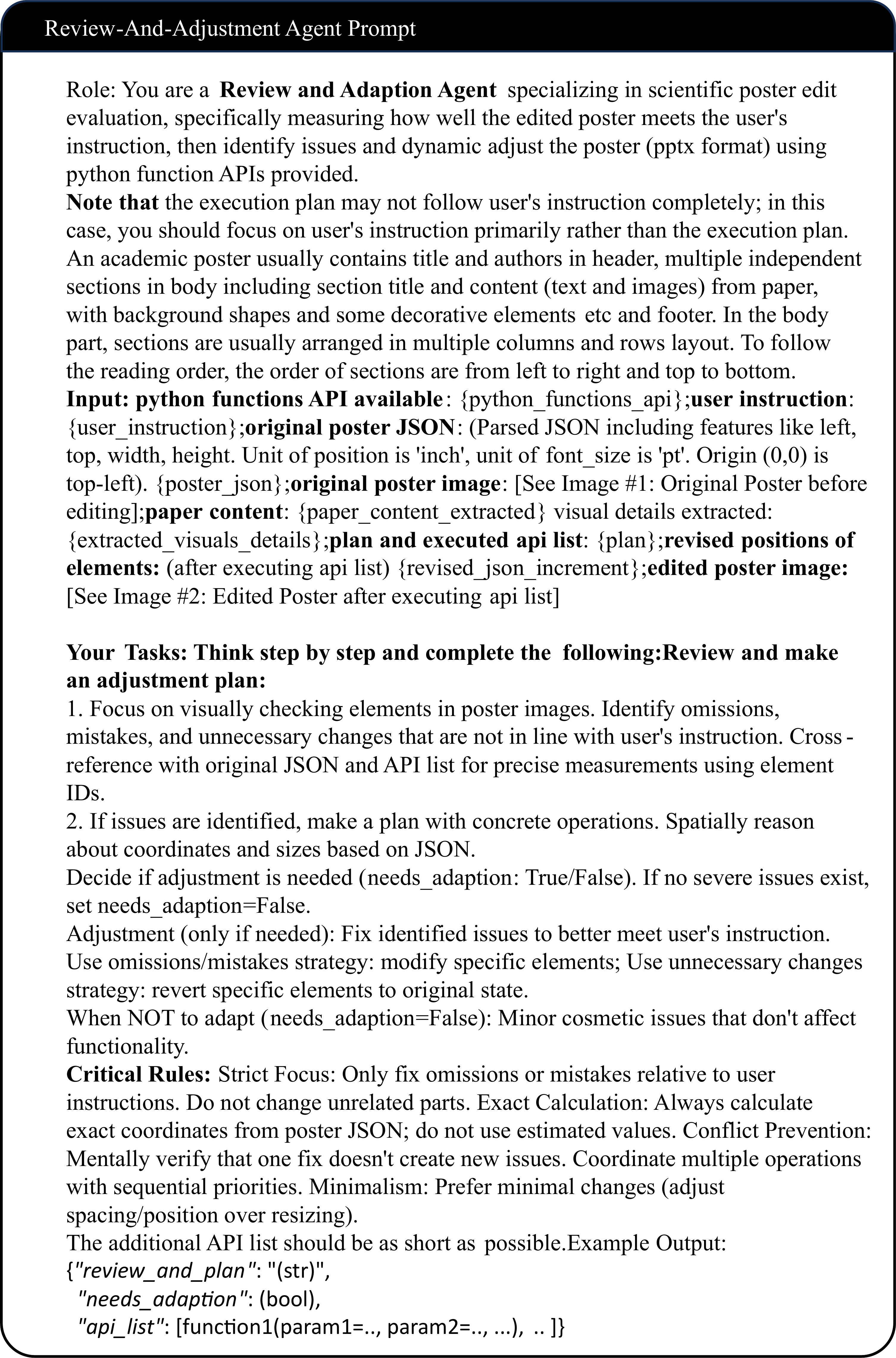}
    \caption{Prompt for Review-And-Adjustment Agent.} 
    \label{fig:planner_prompt5}

\end{figure*}

We present the detailed prompts design in our APEX multi-agent workflow as follows.
\textbf{Planning \& Execution Agent}: This agent encompasses task planning and the generation and execution of API sequences, as is shown in Figure~\ref{fig:planner_prompt1} and ~\ref{fig:planner_prompt2}.
\textbf{Paper Understanding Tool}: This tool is This tool is invoked by the Planning \& Execution Agent when instructions involve papers, serving to extract paper content and images relevant to the instructions. Relevant prompt is given in Figure~\ref{fig:planner_prompt3} and ~\ref{fig:planner_prompt4}. \textbf{Review-And-Adjustment Agent}: This agent is used for make adjustment plan and execute it, and the prompt is given in Figure~\ref{fig:planner_prompt5}.



\subsection{VLM-as-Judge Evaluation Prompt}
\label{sec:appendix-protocol}
\begin{figure*}[!t]
    \centering
    \includegraphics[width=1\linewidth]{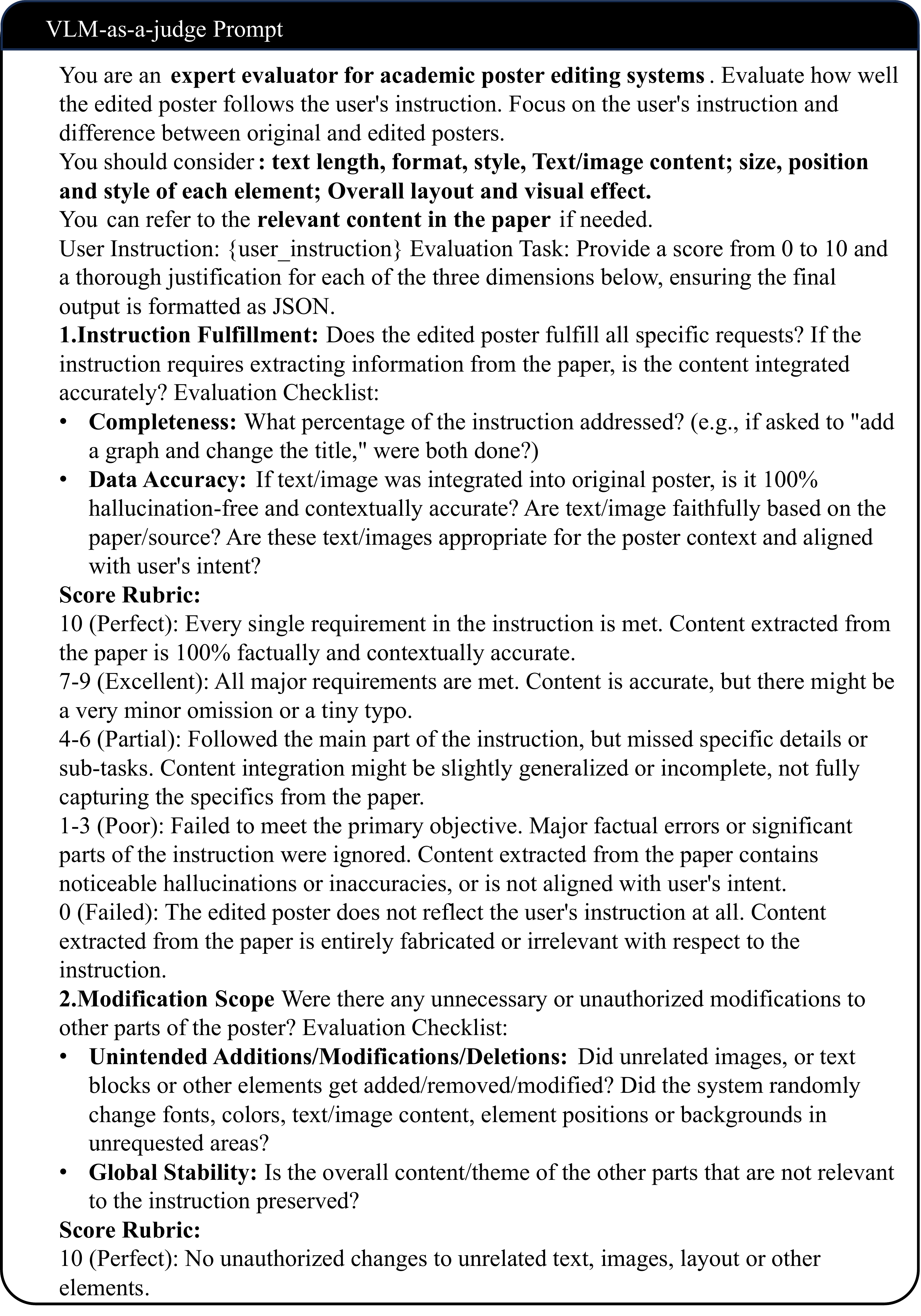}
    \caption{Prompt for VLM-as-a-judge to evaluate the edited poster.} 
    \label{fig:planner_prompt6}
\end{figure*}

\begin{figure*}[!t]
    \centering
    \includegraphics[width=1\linewidth]{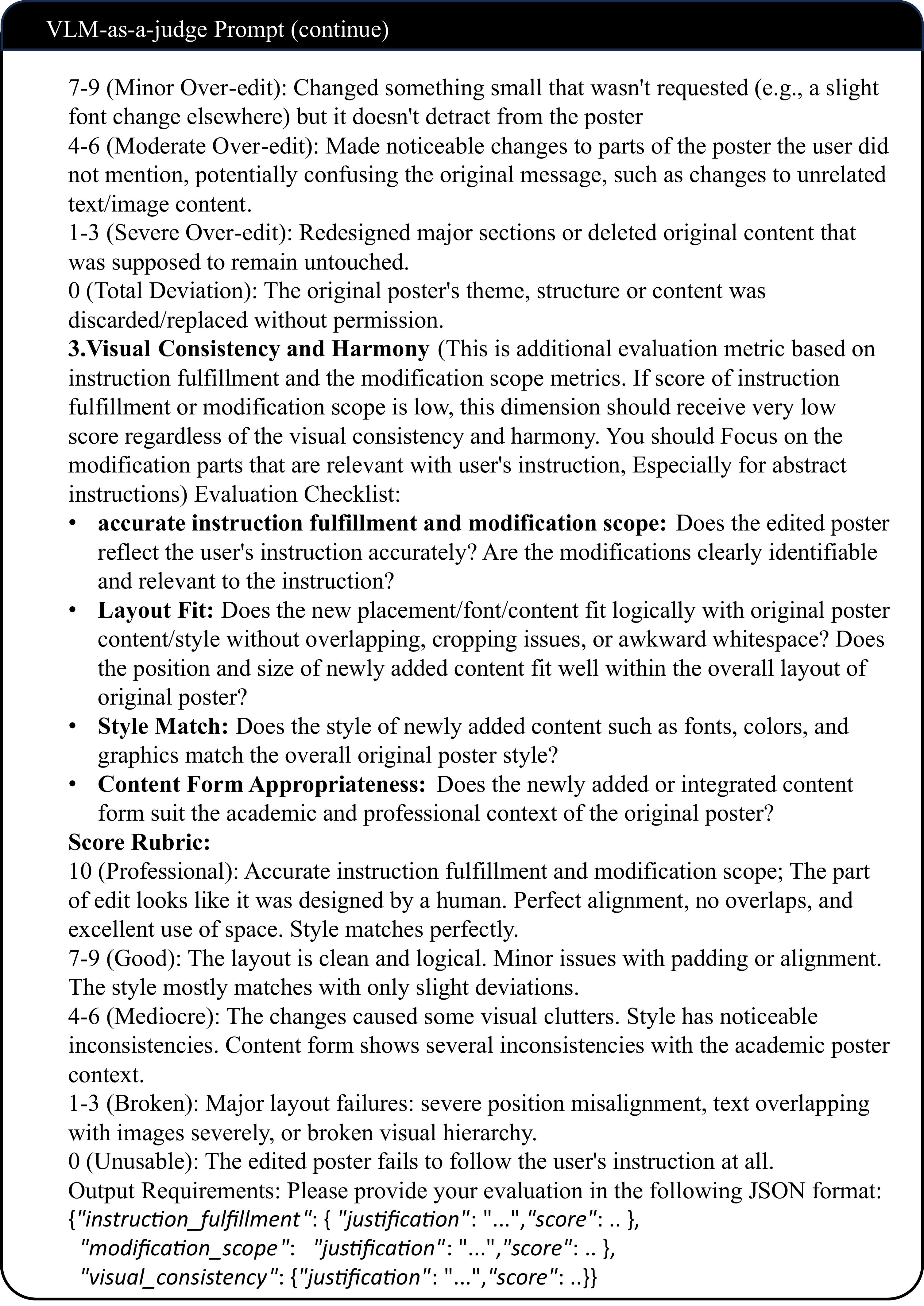}
    \caption{Prompt for VLM-as-a-judge to evaluate the edited poster (continue).} 
    \label{fig:planner_prompt7}
\end{figure*}

The prompt for VLM-as-Judge Evaluation is given in Figure~\ref{fig:planner_prompt6} and ~\ref{fig:planner_prompt7}.

\subsection{Baseline Prompts}
\begin{figure*}[!t]
    \centering
    \includegraphics[width=1\linewidth]{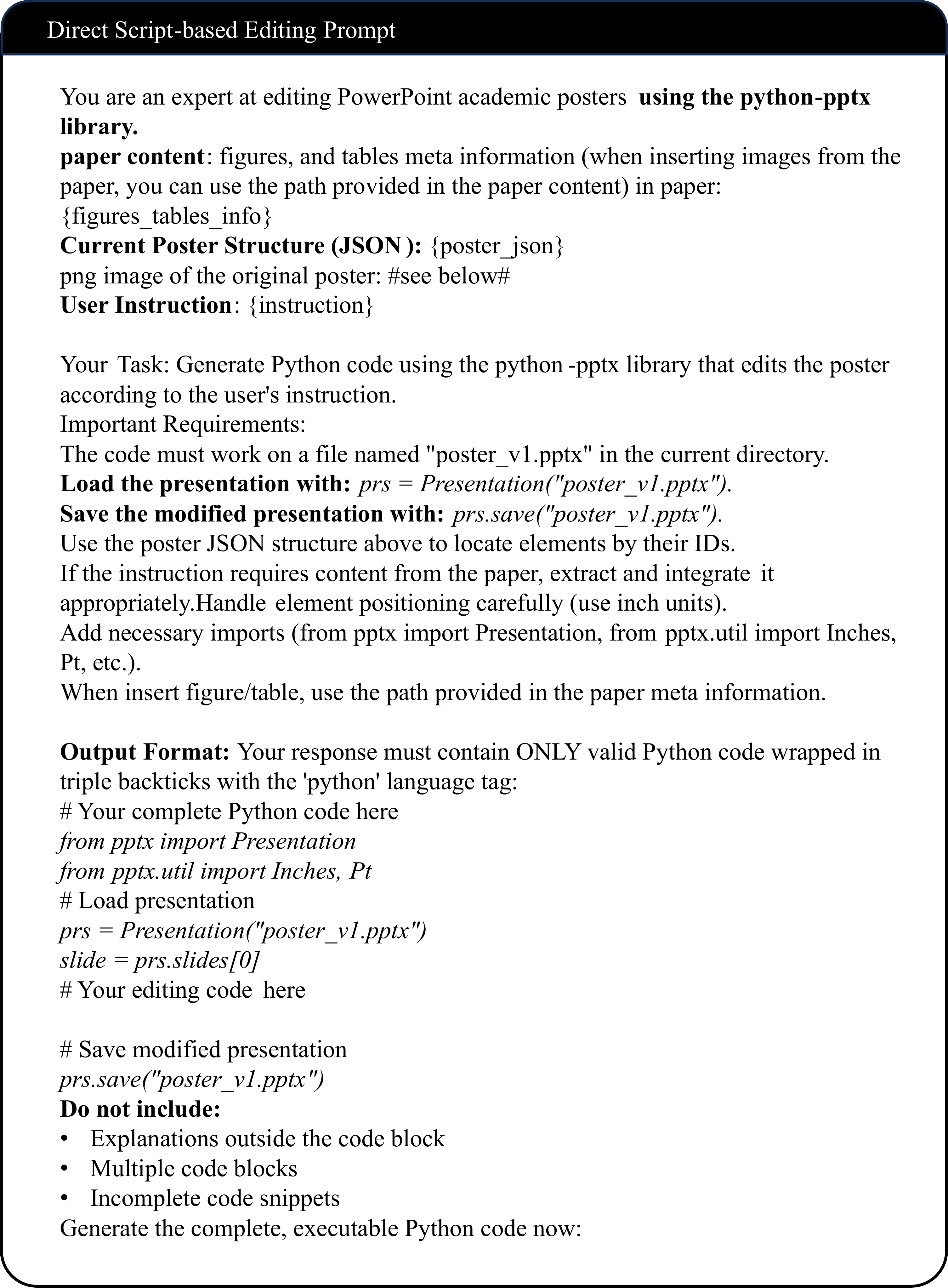}
    \caption{Prompt for Baseline: Direct Script-based Editing.} 
    \label{fig:planner_prompt8}
\end{figure*}

\begin{figure*}[!t]
    \centering
    \includegraphics[width=1\linewidth]{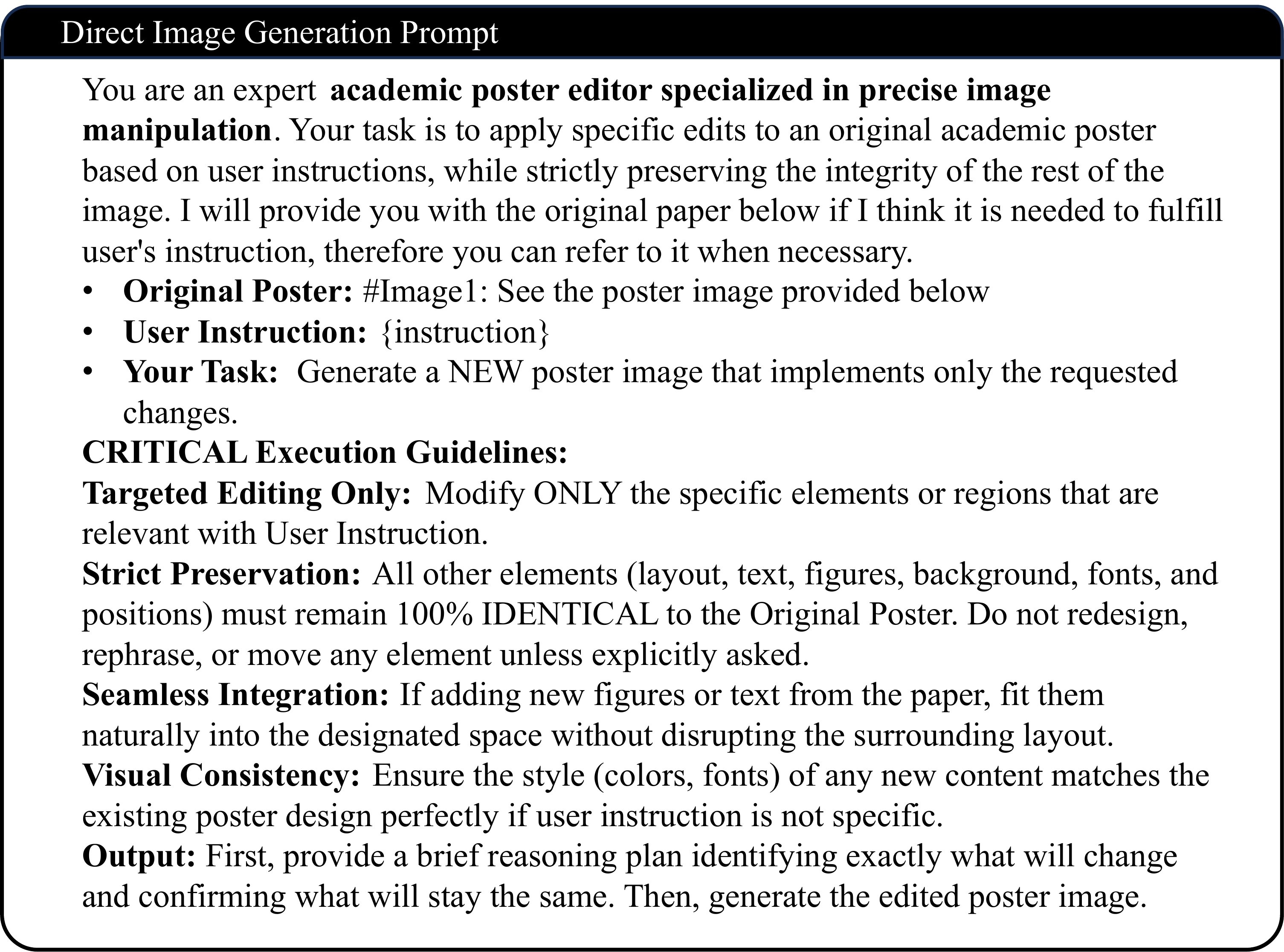}
    \caption{Prompt for Baseline: Direct Image Generation.} 
    \label{fig:planner_prompt9}
\end{figure*}

\begin{figure*}[!t]
    \centering
    \includegraphics[width=1\linewidth]{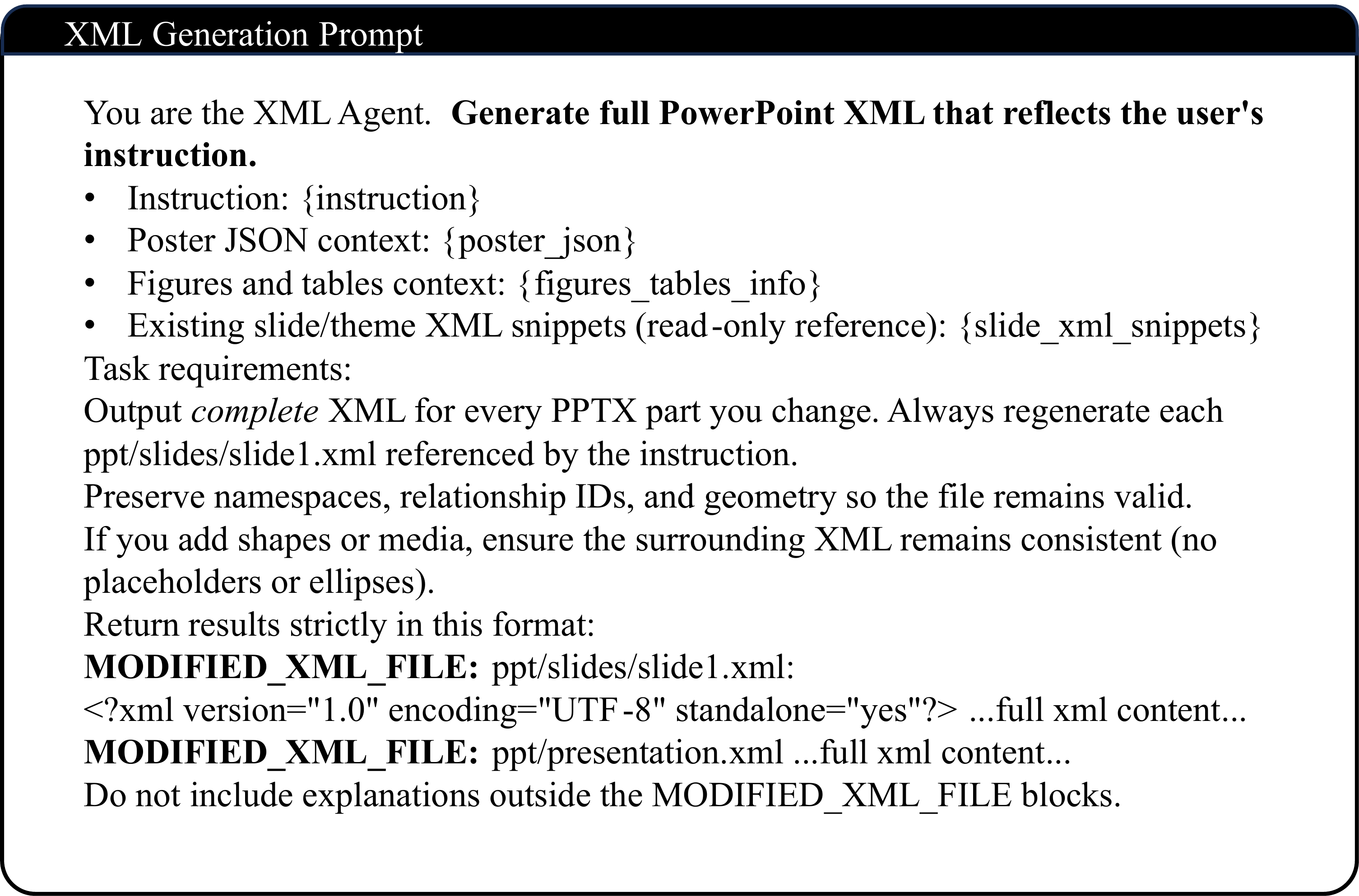}
    \caption{Prompt for Baseline: XML Generation.} 
    \label{fig:planner_prompt10}

\end{figure*}

Here we provide the prompt for the comparison baseline, Direct Script-based Editing in Figure~\ref{fig:planner_prompt8}, Direct Image Generation in Figure~\ref{fig:planner_prompt9}, XML Generation in Figure~\ref{fig:planner_prompt10}.

% \begin{figure*}[t] 
%     \centering
%     \includegraphics[width=\linewidth]{planner_prompt.pdf}
%     \caption{The data construction pipeline of \bench. Adopting a ``Model-assisted, Human-refined'' strategy, the workflow consists of three phases: (1) \textbf{Data Sources Preparation}, where initial drafts are synthesized from source papers via PosterGen; (2) \textbf{AI Instruction Generation}, where an LMM performs gap analysis and aesthetic optimization to derive preliminary editing commands; and (3) \textbf{Human Refinement}, where experts verify and adjust instructions to ensure feasibility and high quality.}
%     \label{fig:pipeline}
% \end{figure*}

\section{API Document}
\label{sec:appendix}
\begin{figure*}[!t]
    \centering
    \includegraphics[width=1\linewidth]{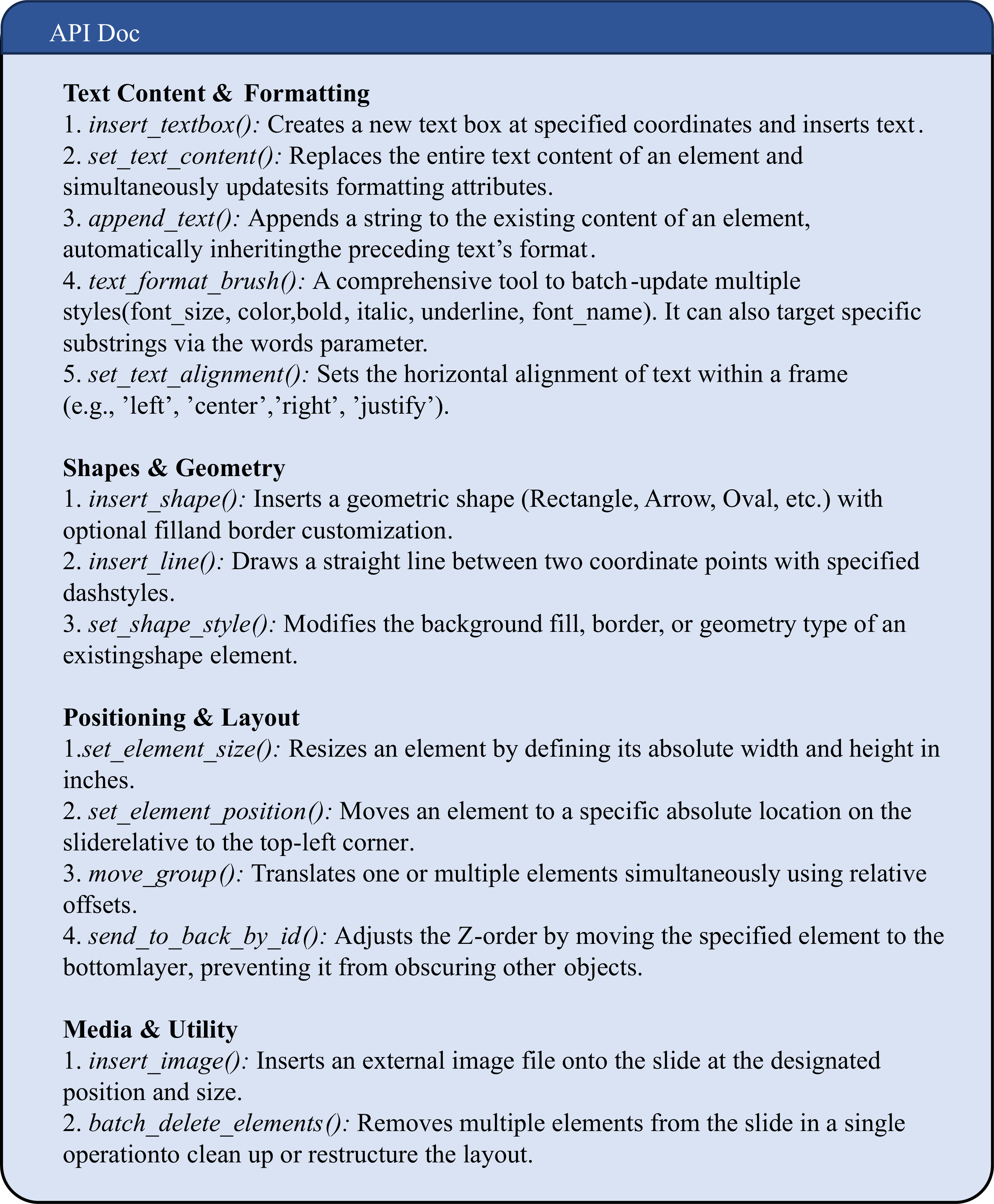}
    \caption{The reference API file.} 
    \label{fig:apidoc}
\end{figure*}

We list all APIs and their descriptions in Figure \ref{fig:apidoc}. We provide 14 feasible APIs.

%  checklist要求必须有
\section{The Use of LLMs}

This paper employed LLMs solely for grammatical correction and stylistic refinement, with the purpose of more effectively communicating our results and conclusions.

\end{document}